\documentclass[a4paper,fleqn]{cas-dc}

\usepackage[authoryear]{natbib}

\usepackage{stfloats}
\usepackage{float} 
\usepackage{booktabs}
\usepackage{tabularx}
\usepackage{caption}
\usepackage{subcaption}
\def\tsc#1{\csdef{#1}{\textsc{\lowercase{#1}}\xspace}}
\tsc{WGM}
\tsc{QE}
\tsc{EP}
\tsc{PMS}
\tsc{BEC}
\tsc{DE}
\begin{document}
% Disable empty ORCID line in the CAS front-matter footnotes when no ORCID IDs are provided.
\RenewDocumentCommand \printorcid {} {}
\let\WriteBookmarks\relax
\def\floatpagepagefraction{1}
\def\textpagefraction{.001}

\shorttitle{Vision–POI Fusion for Building-Level Housing Inspection}

\shortauthors{Zhao et~al.}

\title [mode = title]{How Does Urban Context Relate to Residential Building Health? A Vision–POI Fusion Framework for Building-Level Housing Inspection}                      

\author[1,2]{Kun Zhao}

\author[1,2]{Helei Ren}

\author[1,2]{Guilin Tang}

\author[5]{Tianyi Chen}

\author[1,2]{Zhehui Song}

\author[1,2]{Xing Liu}

\author[1,2]{Lijian Zhou}

\author[3]{Yuhong Zhao}

\author[1,2]{Xiang Gao}

\author[3,4]{Jinming Jiang}

\author[3,4]{Qichao Ban}
\cormark[1]

\affiliation[1]{organization={School of Information Management, Qingdao University of Technology},
    city={Qingdao},
    postcode={266520},
    state={Shandong},
    country={P. R. China}}

\affiliation[2]{organization={Embodied AI \& Robot Research Institute, Qingdao University of Technology},
    city={Qingdao},
    postcode={266520},
    state={Shandong},
    country={P. R. China}}

\affiliation[3]{organization={College of Architecture and Urban Planning, Qingdao University of Technology},
    city={Qingdao},
    postcode={266520},
    country={P. R. China}}

\affiliation[4]{organization={Innovation Institute for Sustainable Maritime Architecture Research and Technology (iSMART), Qingdao University of Technology},
    city={Qingdao},
    postcode={266033},
    country={P. R. China}}

\affiliation[5]{organization={Department of Urban Planning and Design, Xi'an Jiaotong-Liverpool University},
    city={Suzhou},
    postcode={215123},
    country={P. R. China}}

\cortext[cor1]{\begin{tabular}[t]{@{}l@{}}Corresponding author.\\ E-mail address: qichao.ban@qut.edu.cn (Q. Ban).\end{tabular}}

\begin{abstract}
Housing-level urban physical examination is essential for identifying residential building problems and supporting targeted urban renewal.
Existing automated inspection studies, however, primarily rely on individual images and rarely examine whether surrounding urban functional context can provide supplementary information for building-level assessment.
This study proposes a vision--point-of-interest (POI) fusion framework that combines multi-view visual inspection with POI-derived neighborhood context for residential building health assessment.
The empirical dataset covers 92 old residential communities, 3,237 residential buildings, and 25,608 field-acquired inspection images in Qingdao, China, encompassing seven categories of housing-related issues.
First, multiple object detection models are evaluated to extract issue locations, categories, and confidence scores from individual images.
The image-level outputs are subsequently aggregated across multiple views to construct interpretable building-level representations, including detection frequency, confidence statistics, and view-level occurrence rates.
Second, POI features are extracted within 500~m, 1,000~m, and 1,500~m neighborhood buffers to characterize surrounding functional environments.
Pearson and Spearman correlation analyses, combined with false discovery rate correction, are used to identify candidate contextual features.
Finally, visual and POI features are integrated using a cost-sensitive Random Forest classifier under community-isolated spatial cross-validation.
The results show that multi-view aggregation provides the main performance improvement, increasing the building-level Macro-F1 from 60.84\% under Direct Detection to 74.95\% under Multi-View Visual Aggregation.
Incorporating POI context further increases Macro-F1 to 76.79\%, although the additional gain is modest and category-dependent.
POI information therefore functions as a supplementary contextual prior rather than a substitute for direct visual evidence or a causal determinant of building condition.
The proposed framework demonstrates how visual inspection and urban spatial data can be jointly used to support scalable building-level housing assessment and refined urban renewal governance.
\end{abstract}

% \includegraphics{figs/grabs.pdf}

%\item Multi-view images and multi-scale points of interest (POIs) form a building-linked inspection dataset.
%\item Vision–POI fusion supports contextual post-correction at the building level.
%\item Community-isolated validation tests within-city generalization to unseen communities.

% Keywords

\maketitle

\section{Introduction}
Urban renewal has increasingly been recognized as a critical pathway to realizing high-quality urban development and sustainable urban governance worldwide, acting as a fundamental strategy to address functional decline, environmental degradation, infrastructure aging, and spatial inequality in established urban areas.
Rather than relying exclusively on large-scale redevelopment, contemporary renewal policies increasingly emphasize systematic diagnosis, targeted intervention, and the continuous improvement of existing residential environments \citep{HUANG2020102074}.
As a scientific evidence base for this process, China's "urban physical examination" framework evaluates urban development performance across a multi-dimensional index system encompassing spatial structure, infrastructure, environmental quality, and public services \citep{CHEN2022108602}.
The Chinese government has officially defined a four-level architecture—housing, community, neighborhood, and city \citep{mohurd2023guiding}.
Among these, the housing-level comprehensively reflects the overall quality and livability of residential buildings, typically encompassing assessments of building structural safety, damage to building components, occupation of public spaces, operational efficiency of supporting facilities, and accessibility retrofitting for aging populations—which are precisely the core concerns of people-centered urban renewal.

Unlike assessments at other levels that primarily evaluate the exterior built environment via remote sensing or street-view imagery, housing-level inspection focuses mainly on interior building conditions (e.g., entrances, corridors, stairwells, and basements).
These semi-enclosed spaces remain invisible from aerial or street-level perspectives, making comprehensive large-scale data collection difficult using unmanned aerial vehicles (UAVs) or mobile mapping systems alone.
Consequently, existing practices remain heavily dependent on labor-intensive field surveys and expert judgment, resulting in limited scalability, high labor costs, and inconsistent evaluation standards.
This further leads to a severe scarcity of dedicated, building-linked image datasets for training intelligent inspection models.

Among intelligent inspection technologies, computer vision (CV) has rapidly advanced in recent years.
It has been widely applied in various urban contexts~\citep{MARASINGHE2026107209},
including but not limited to building change detection and unauthorized building monitoring~\citep{9311793,10476617,11078759,LIU2026113052},
building facade damage detection~\citep{s19163556,WANG2023104810,buildings13112754,10747365,su17219390,s25237118,11129785,buildings15111865,s26020694,11398095},
crack detection~\citep{nguyen2023deep,buildings13071814,buildings14071929,GE2025105951,wang2025building,gamage2026event},
and post-disaster building damage assessment~\citep{9924933,xia2023deep,10683869,10747526,rs17243957}.

Regarding building change detection and unauthorized construction monitoring, \cite{9311793} designed a Siamese network with a dual attention mechanism and improved Focal Loss function, achieving high-precision building change detection in remote sensing imagery.
~\cite{10476617} proposed IBMNet, an edge-focused Siamese network, for illegal building detection in satellite imagery. 
Furthermore,~\cite{LIU2026113052} developed IBDNet using high-tower cameras for real-time detection and localization of illegal buildings, meeting the requirements for illegal construction screening in community-level residential surveys.
In building facade damage detection,
~\cite{WANG2023104810} proposed a 2D/3D vision-coupling method, improving ResNet-50 and Grad-CAM to integrate defect recognition with 3D reconstruction.
~\cite{buildings13112754} developed drone inspection schemes and detection models tailored to residential privacy concerns and facade complexity.
~\cite{buildings15111865} incorporated normalization-based attention mechanism (NAM) modules into YOLOv8 to address the issue of missed detections in fine cracks.
Regarding the optimization of crack detection,
~\cite{buildings14071929} proposed a hybrid data augmentation strategy combining 3D virtual generation with real-image fusion to address the problem of insufficient training data. 
~\cite{GE2025105951} developed a streamlined ground-robot platform integrating LiDAR and monocular cameras, achieving automatic segmentation and 3D visualization of structural damage.
~\cite{gamage2026event} constructed the first event-based civil infrastructure defect dataset, ev-CIVIL, overcoming performance bottlenecks of traditional cameras under extreme lighting conditions.
In the domain of post-disaster damage assessment,
~\cite{xia2023deep} integrated the two-stage BDANet with ultra-high-resolution remote sensing to achieve rapid and precise assessment of building damage in earthquake-affected areas.
~\cite{10747526} combined the SLgViT network with weakly supervised learning, utilizing bi-temporal remote sensing images for pixel-level building damage detection.
~\cite{rs17243957}  decoupled building localization and damage classification, employing YOLO for localization and Transformer for classification, achieving superior performance compared to traditional joint models.

Visual data are frequently integrated with Geographic Information System (GIS) data to construct multimodal datasets, collaboratively fulfilling comprehensive building defect inspection tasks~\citep{CHEN2021103503,HUANG2026106435}.
%In terms of GIS fusion,
~\cite{CHEN2021103503} developed a framework for georegistering UAV imagery to GIS spatial models;~\cite{HUANG2026106435} proposed urban visual intelligence methods, achieving large-scale mapping and quantification of building damage at urban scales.

Furthermore, intelligent vision technology has been widely utilized to provide objective evidence for the livability assessment of community life circles.
In recent years, vision technologies (e.g., street view imagery, UAV, remote sensing) have become core tools for quantifying 15-minute community life circles.
~\cite{LI2023138883} utilized remote sensing to analyze illegal land use patterns within ecological spaces,
while~\cite{rs15112845} systematically reviewed the application potential of UAVs across multiple urban planning scenarios.
The latest research further integrates multimodal data with advanced algorithms to analyze in a deeper way the impact of urban morphology on life circle vitality~\citep{HE2024103287}
and~\cite{LIN2025102246} used multi-layer POI networks to examine interactions among street-level urban functions.

Based on the current research status of the application of intelligent visual technology in urban physical examinations, the following observations can be made.
Despite the widespread adoption of various intelligent technologies in urban assessments ranging from the community to the neighborhood scale, there remains a notable absence of large-scale, housing-level urban physical examinations conducted within the interior of buildings.
In addition, although intelligent vision technologies have been applied in neighborhood livability assessments represented by the 15-minute city concept, they are generally treated merely as tools for neighborhood evaluation.
Whether neighborhood functional context provides supplementary predictive information for building-level housing inspection remains insufficiently examined.

In this context, this study develops a vision–POI fusion framework for housing-level urban physical examination, covering interior common spaces, semi-enclosed areas, and visible exterior building problems.
The main contributions of this study are summarized as follows:
\begin{itemize} 
\item 
   First, this study constructs a building-linked housing inspection dataset covering 92 old residential communities, 3,237 residential buildings, and 25,608 field-acquired images. Seven categories of housing-related issues are defined, and POI data within 500 m, 1,000 m, and 1,500 m neighborhood buffers are spatially linked to the visual records, supporting integrated analysis at the image, building, and community levels.
\item    
    Second, this study proposes a building-level vision–POI fusion framework combining multi-view visual aggregation with contextual post-correction. Image-level detections are aggregated into interpretable building-level visual features, while multi-scale POI indicators are introduced as supplementary contextual priors to support the correction of missed detections and false positives. POI information is treated as predictive context rather than as a causal determinant of building condition.
\item
    Finally, the framework is systematically evaluated across 92 representative old residential communities in Qingdao using multiple detectors, neighborhood scales, and spatially isolated validation settings. The results show that multi-view aggregation provides the primary improvement in building-level inspection, while POI context yields an additional and category-dependent gain, demonstrating the practical potential of visual–GIS integration for large-scale housing assessment and refined urban renewal governance.
\end{itemize}

\begin{figure*}[b]
  \centering
  \includegraphics[width=\textwidth]{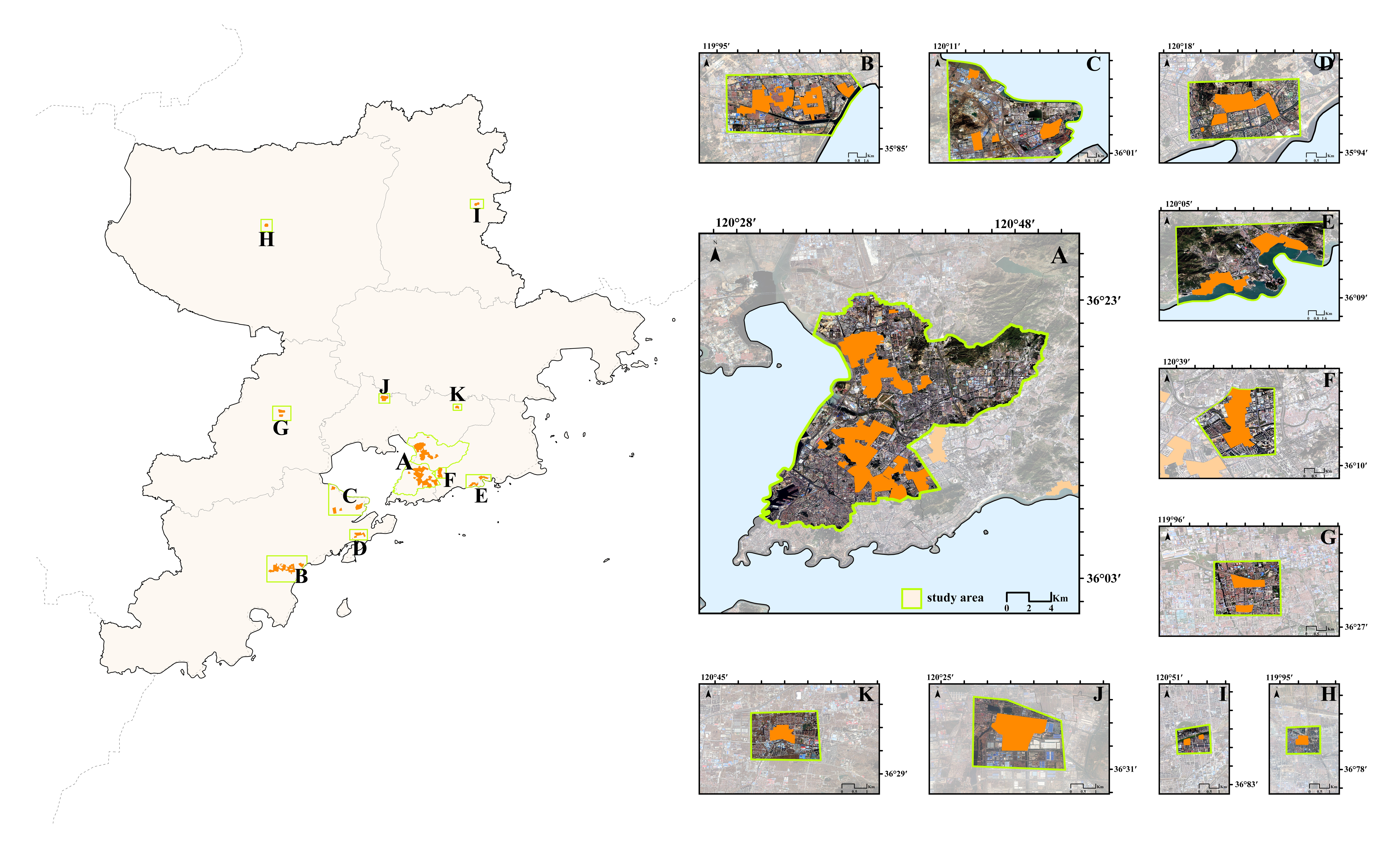} 
  \caption{Spatial distribution of the 92 sampled residential communities in Qingdao. The main map shows the administrative boundary of Qingdao and the locations of 11 spatial clusters of old residential communities (A--K), highlighted in orange.}
  \label{fig:study_area}
\end{figure*}
\section{Study Area and Data}
\subsection{Study Area Overview}
This study selects Qingdao, Shandong Province, China, as the study area. Qingdao is a major coastal port city with a substantial stock of residential communities constructed during the rapid urban expansion of the 1980s and 1990s.
After decades of use, many of these communities face common problems, including aging building components, outdated supporting facilities, insufficient accessibility improvements, and uneven property management.
Unauthorized alterations and additions are also observed in some communities, making these areas important targets for urban renewal and refined residential governance.

Qingdao was selected for two main reasons.
First, the availability of field inspection records and spatial information provided a suitable basis for conducting systematic image acquisition and building-level data linkage across multiple communities.
Second, Qingdao has continuously implemented old residential community renovation and urban physical examination programs in recent years, providing a relevant policy and practical context for evaluating data-driven housing inspection methods.

The original inspection records covered multiple administrative districts and included 15 categories of housing-related issues.
Following the data cleaning, category integration, and building-matching procedures described in Section~\ref{section2.2}, 92 old residential communities were retained as the empirical study sample.
These communities differ in construction period, building scale, spatial location, and management conditions, thereby representing a range of typical old residential environments in Qingdao.
Their spatial distribution is shown in Figure~\ref{fig:study_area}.

\subsection{Problem Definition and Data Preprocessing}
\label{section2.2}
Based on policy documents and national standards issued by the Ministry of Housing and Urban-Rural Development of China~\citep{GB1, GB2, mohurd2023guiding}, together with the characteristics of old residential buildings in Qingdao, 15 original categories of housing-related issues were defined during field data collection.
The original categories are listed in the middle column of Table~\ref{tbl:category_mapping}.
Because several categories contained relatively few valid samples, semantically related categories were subsequently consolidated into seven core categories for model training and statistical analysis.

The category integration followed two principles.
First, semantic consistency was considered by merging categories that described similar problems or belonged to the same building component or functional space.
Second, sample adequacy was considered by combining low-frequency categories with visually and semantically related categories to ensure that each final category contained sufficient samples for model training.
Based on these principles, the original 15 categories were mapped to seven core categories. The detailed mapping relationships and integration rationale are presented in
Table~\ref{tbl:category_mapping}.
\begin{table*}[ht]
\caption{Mapping of the 15 original housing inspection categories to the 7 core categories used in this study.}
\label{tbl:category_mapping}
\renewcommand{\arraystretch}{1.3}
\begin{tabularx}{\linewidth}{@{} l X X @{}}
\toprule
\textbf{Final Category (7 Categories)} & \textbf{Original Categories (15 Categories)} & \textbf{Integration Logic and Description} \\
\midrule
illegal renovation expansion & 1. Illegal demolition/alteration, damage to load-bearing structures, adding floors, expansion, etc. \newline 2. Increased floor (roof) live loads & Merges illegal demolition/alteration behaviors with the resulting increased loads to strengthen the recognition of illegal reconstruction and its consequences. \\

addition balconies windows & 1. Illegal addition of bay windows or balconies & This category has distinct features and a moderate sample size, so it is retained as an independent category. \\

public spaces illegal occupation & 1. Safety hazards in corridor facilities \newline 2. Insufficient lighting at unit entrances \newline 3. Covered or damaged smoke detectors in public areas of high-rise residences & Merges management deficiencies or safety hazards in public areas such as corridors, entrances, and public fire facilities, collectively termed public space issues. \\

wall damage & 1. Damage, cracks, or deformation of load-bearing components (stairs) \newline 2. Cracking, damage, or falling of exterior wall decoration and insulation materials \newline 3. Safety hazards in enclosure facilities \newline 4. Obvious instability, sliding, or settlement of building foundations \newline 5. Backflow hazards in residential basements \newline 6. Safety hazards in open-air parts & Merges various physical damages involving the main building structure, enclosure systems, and foundations into a broad "wall/structural damage" category. \\

elevator addition & 1. Safety hazards in elevators & Groups elevator-related safety hazards here, covering common issues in old communities such as the addition of elevators or lack of maintenance. \\

pipeline damage & 1. Damage, blockage, or leakage of water supply and drainage pipelines & This category has strong independence and is retained directly. \\

no aging place modifications & 1. Issues with aging-in-place modifications & This category has strong independence and is retained directly. \\
\bottomrule
\end{tabularx}
\end{table*}

Following the 2025 Qingdao Urban Physical Examination Work Plan implemented by the Qingdao Municipal Housing and Urban-Rural Development Bureau, the inspection records were further cleaned and linked to individual residential buildings.
The resulting building-level empirical dataset covers 92 old residential communities and 3,237 residential buildings.

Several data cleaning procedures were applied.
First, records with incomplete information or missing key attributes were removed.
Second, images that could not be reliably linked to a specific residential building, as well as images with insufficient visual quality for inspection, were excluded.
Third, communities with very large or very small numbers of sampled buildings were spatially reorganized during data management to improve the balance of community-level analytical units.
Large residential areas were divided into smaller spatial units where necessary, whereas geographically adjacent micro-communities with similar spatial and management characteristics were combined.

\begin{figure*}[t]
  \centering
  \includegraphics[width=\textwidth]{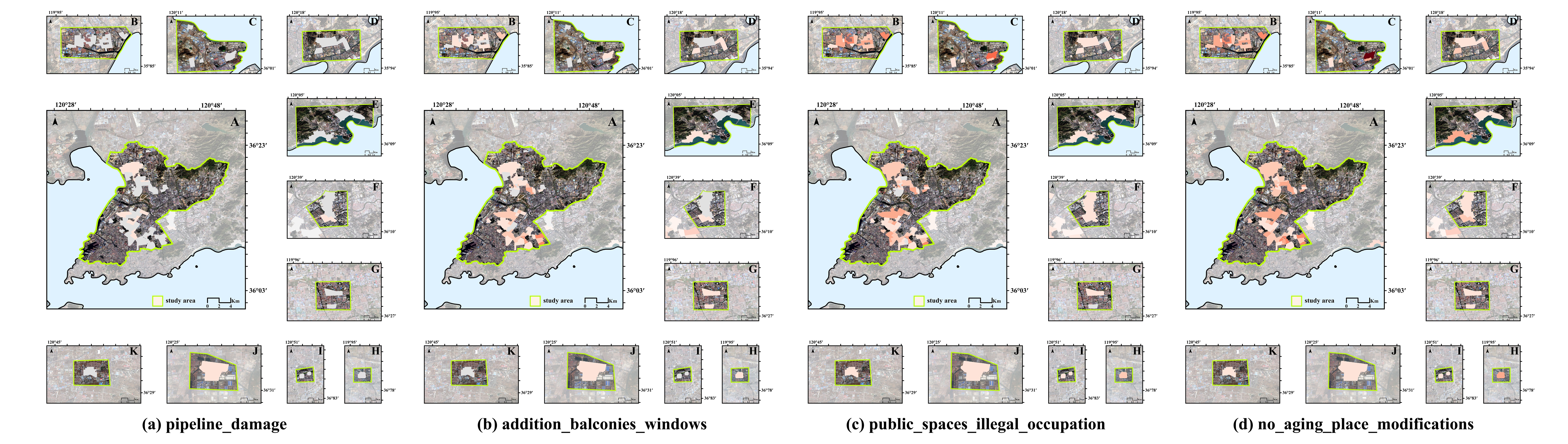}
  \vspace{0.7cm} 
  \includegraphics[width=\textwidth]{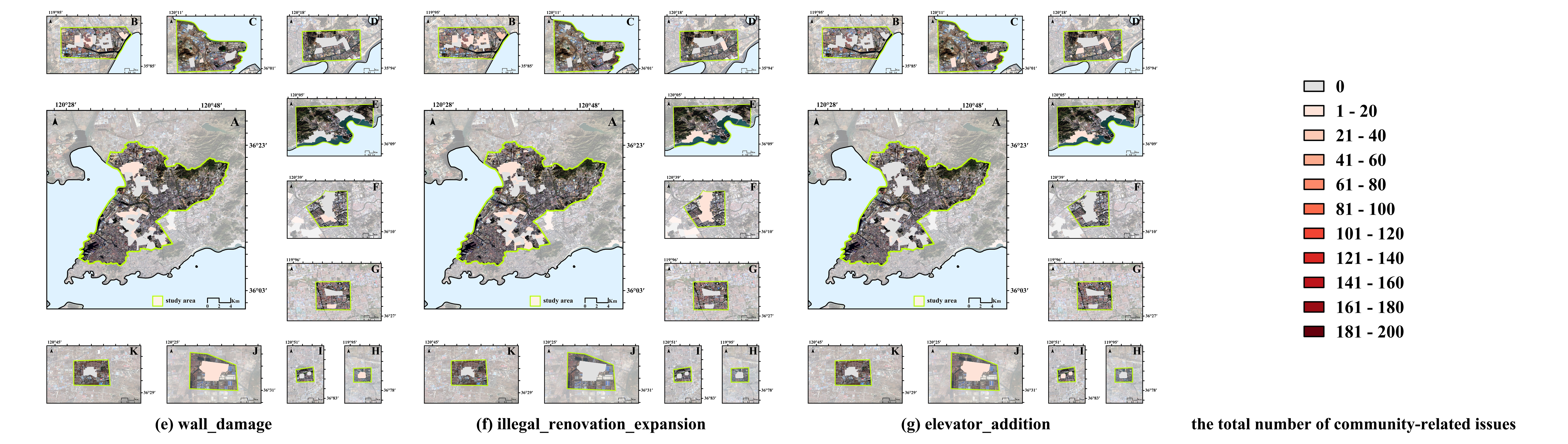}
  \caption{Spatial distribution of the seven categories of housing-related issues across the 92 sampled communities in Qingdao. Color intensity represents the number of recorded issue instances in each community.}
  \label{fig:issue_distribution}
\end{figure*}

After data cleaning and screening, the final empirical dataset covered 92 old residential communities across eight administrative districts and 27 subdistricts in Qingdao, including 3,237 residential buildings and 25,608 multi-view inspection images.
Each building was associated with an average of 7.91 images, with a median of seven images.
Figure~\ref{fig:issue_distribution} presents the spatial distribution of the seven categories of housing-related issues across the sampled communities.
The variation in color intensity indicates the number of recorded issue instances in each community.
The results show clear spatial heterogeneity, with larger sample concentrations in Shibei District, the West Coast New Area, and Licang District, and smaller sample concentrations in Pingdu, Jiaozhou, Chengyang, and Laixi.

\subsection{Life Circle and POI Data}
The POI data for this study was retrieved via the AutoNavi Maps Open Platform API.
AutoNavi is a widely used digital map service in China and provides detailed POI records covering commercial services, public facilities, educational institutions, healthcare services, accommodation, and employment-related functions. 
The spatial scope of data collection encompasses the entire study area of Qingdao, allowing the POI records to be spatially linked to the 92 sampled residential communities.

The concept of the 15-minute community life circle was used to define the maximum extent of the surrounding neighborhood context.
For each community, a reference point was calculated as the arithmetic mean of the geographic coordinates of all sampled residential buildings within that community.
Circular POI search buffers were then generated around each community reference point.

A radius of 1,500~m was used as an approximate spatial proxy for the upper bound of a 15-minute neighborhood context.
This Euclidean-distance approximation does not represent an exact pedestrian catchment because road-network connectivity, physical barriers, terrain, and community entrances were not explicitly modeled.
To examine the sensitivity of contextual information to spatial scale, three buffer radii of 500~m, 1,000~m, and 1,500~m were constructed for each community.

Based on the primary POI classification system of AutoNavi Maps and the objectives of this study, the retained POIs were grouped into seven functional categories: dining services, shopping services, accommodation, healthcare, education/culture, life services, and corporate offices.
Dining, shopping, and accommodation POIs primarily characterize commercial and visitor-oriented functions;
healthcare and education/culture POIs represent major public service functions;
and life services and corporate offices reflect daily service provision and employment-related activities.

In terms of data cleaning, records with abnormal coordinates (lying outside the study area), duplicate entries (resulting from multiple collections of the same location), and POI records with remarkably low relevance to the daily functional environment surrounding old communities (such as industrial facilities in remote suburbs) were progressively eliminated through the application of latitude and longitude constraints, category filtering criteria, and multi-round data scrubbing procedures.
Consequently, this study constructed a POI dataset that includes functional facility information related to building inspection issues within the 92 old residential communities.
The 500~m buffers contained approximately 41,713 POI records, the 1,000~m buffers contained 91,403 records, and the 1,500~m buffers contained 111,614 records.
These three spatial scales were subsequently used to examine how the extent of neighborhood context influenced the statistical association analysis and building-level fusion model.

\subsection{Training Data and Annotation}
To construct a robust visual inspection model, this study introduced the \textbf{H}ousing-dimensi\textbf{O}nal vis\textbf{U}al in\textbf{S}pection imag\textbf{E} \textbf{D}ataset (\textbf{HOUSED}), which was previously compiled and publicly released by our research team, as an important source for pre-training and auxiliary training.
The original HOUSED dataset contains 19 fine-grained categories.
To align with the 7-core-category system of this study, it was necessary to filter and map them.
The mapping process followed the principles of semantic correlation and visual feature consistency, merging HOUSED categories with similar semantics or visual features into the 7-category system of this study.
The specific mapping relationships and selection rationale are shown in Table~\ref{tbl:housed_mapping}.
\begin{table*}[ht]
\caption{Mapping and selection rationale between this study’s categories and the HOUSED dataset.}
\label{tbl:housed_mapping}
\renewcommand{\arraystretch}{1.3}
\begin{tabularx}{\linewidth}{@{} l >{\raggedright\arraybackslash}X >{\raggedright\arraybackslash}X @{}}
\toprule
\textbf{Study Category (7 Categories)} & \textbf{HOUSED Original Categories (19 Categories)} & \textbf{Selection and Mapping Rationale} \\
\midrule
illegal renovation expansion & load bearing component damaged, floor roof excessive load, renovation extension unauthorized & Merges structural damage, load increases, and unauthorized modifications directly related to illegal renovation and expansion. \\

addition balconies windows & balcony window addition unauthorized & Directly maps the issue of unauthorized additions of balconies or bay windows. \\

public spaces illegal occupation & public space occupation illegal, electric vehicle corridor charging illegal & Merges illegal occupation of public spaces and unauthorized charging of electric vehicles in corridors. \\

wall damage & facade material damaged, roof floor finish damaged, glass door window damaged & Merges physical damage to the building's enclosure system, including facades, roofs, and doors/windows. \\

elevator addition & elevator installation absence addition & Directly maps the absence of necessary elevator installations or the addition of new elevators. \\

pipeline damage & pipeline water electricity telecom damaged & This category is clearly defined in HOUSED and is retained as a direct mapping. \\

no aging place modifications & handrail guardrail missing damaged, step missing damaged, ramp anti slip entrance corridor absence & Merges the absence or damage of elderly-oriented facilities such as handrails, steps, and non-slip ramps. \\
\bottomrule
\end{tabularx}
\end{table*}
\begin{figure*}[!h]
\centering
  % ---- a ----
  \begin{subfigure}{0.40\textwidth}
    \centering
    \includegraphics[width=\textwidth]{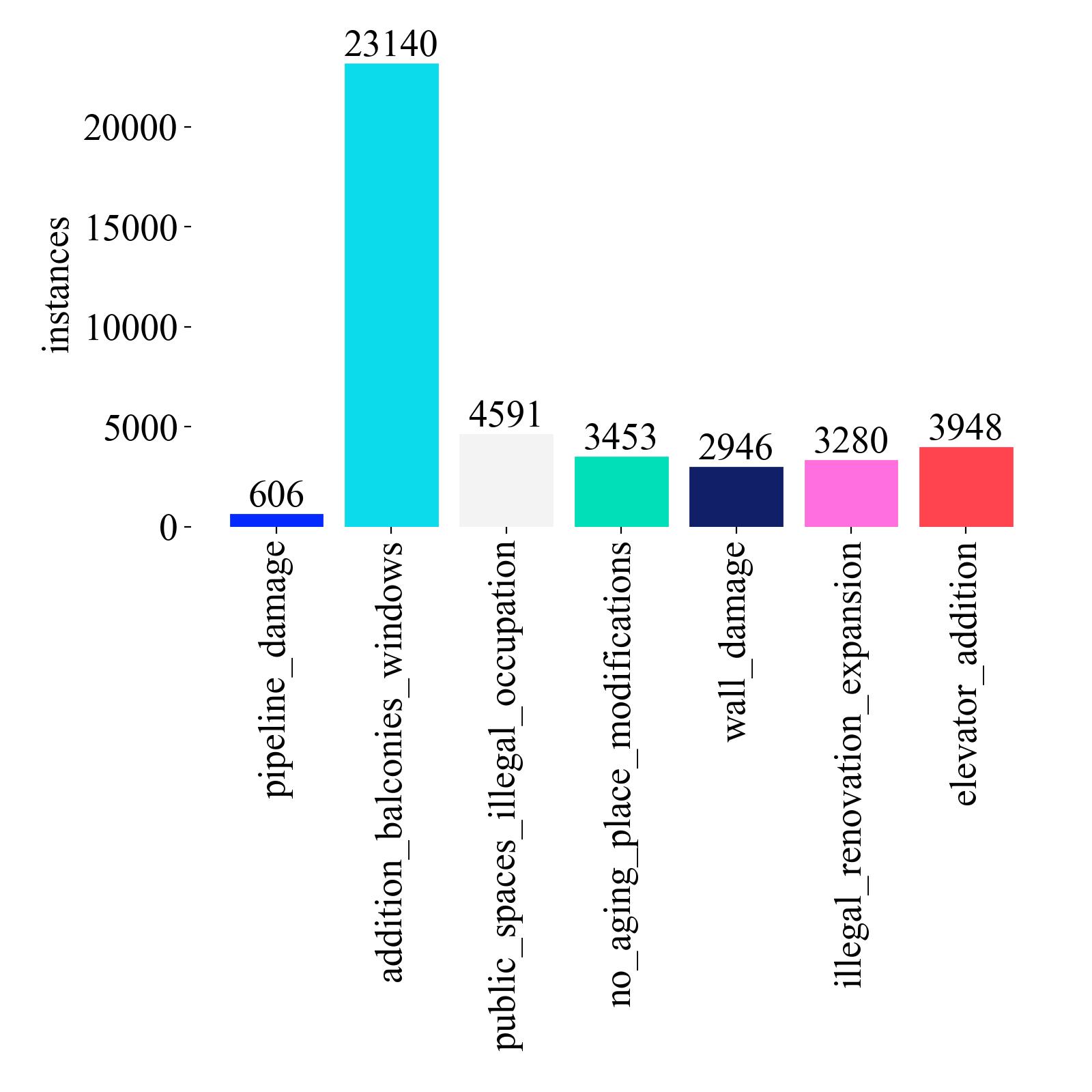}
    \caption{Histogram of category instance distribution.}
    \label{fig3_1}
  \end{subfigure}
  \hfil 
  % ---- b ----
  \begin{subfigure}{0.40\textwidth}
    \centering
    \includegraphics[width=\textwidth]{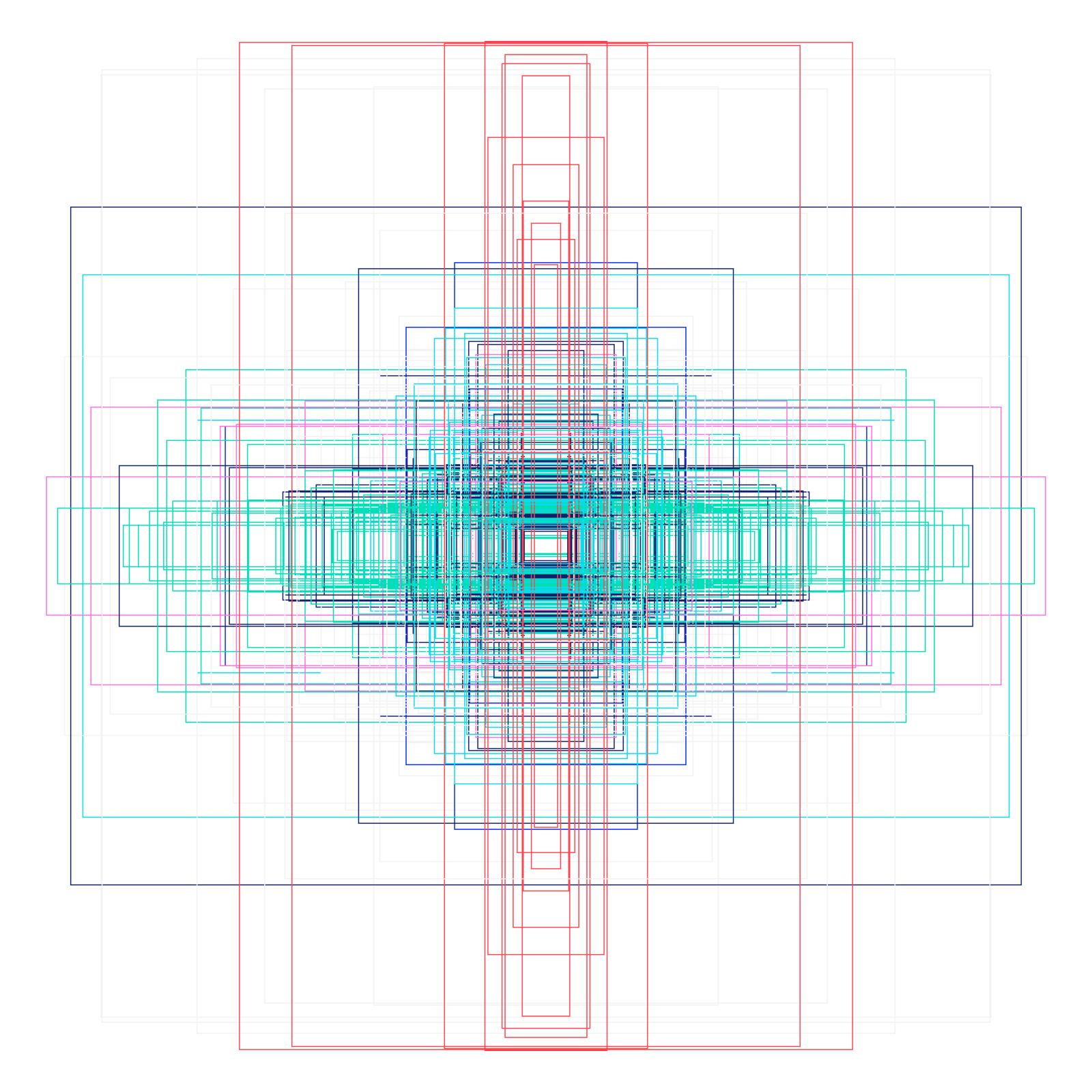}
    \caption{Bounding box spatial overlay.}
    \label{fig3_2}
  \end{subfigure}

  \vspace{1pt}

  % ---- c ----
  \begin{subfigure}{0.40\textwidth}
    \centering
    \includegraphics[width=\textwidth]{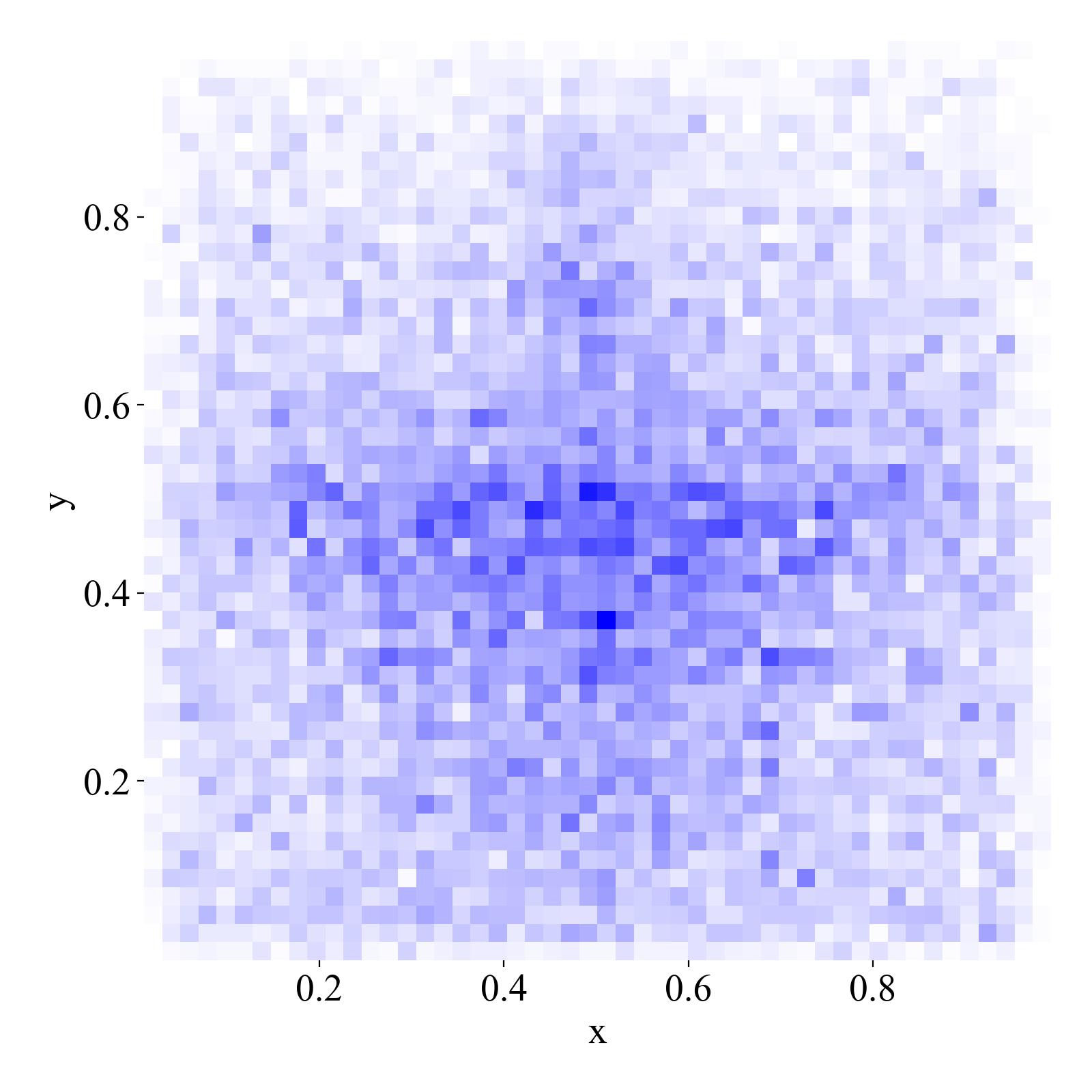}
    \caption{Center point location heatmap.}
    \label{fig3_3}
  \end{subfigure}
  \hfil 
  % ---- d ----
  \begin{subfigure}{0.40\textwidth}
    \centering
    \includegraphics[width=\textwidth]{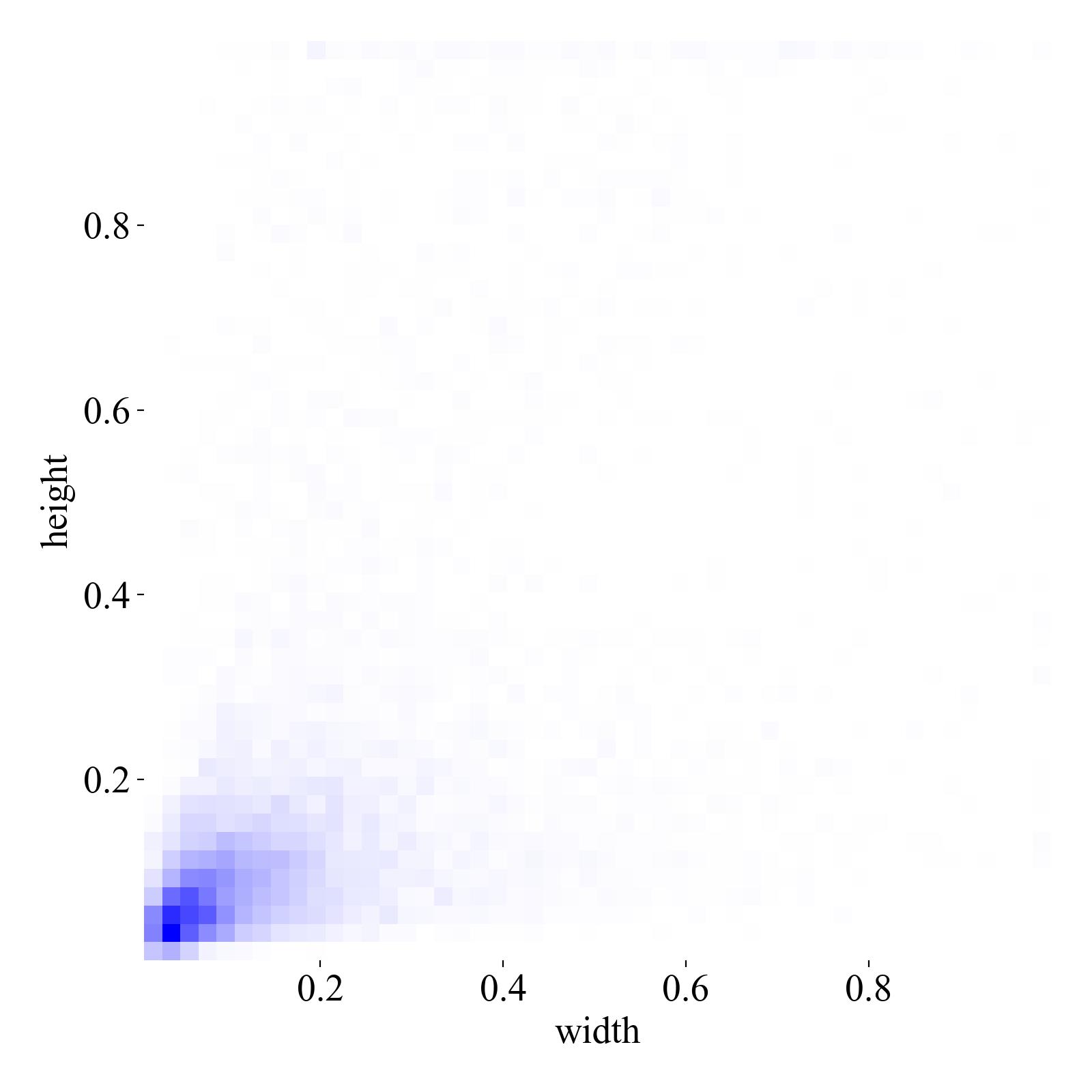}
    \caption{Heatmap of target size distribution.}
    \label{fig3_4}
  \end{subfigure}

\caption{Four-dimensional visual analysis of the finalized balanced training set: (a) displays the number of annotated instances for the 7 core categories; (b) overlays bounding boxes in a normalized space to show geometric forms and common aspect ratios; (c) shows the distribution of center coordinates within the image coordinate system; (d) reflects the relationship between the width and height of annotated bounding boxes.}
\label{fig3_total} 
\end{figure*}

After completing the category mapping, the research team performed rigorous data inspection and annotation correction on the selected HOUSED data to ensure complete consistency in format and standards with the independently collected Qingdao local dataset.
First, samples with blurred images, severe occlusion, or unclear annotations were removed from the HOUSED dataset to ensure the quality of training data.
Based on the mapping relationships in Table~\ref{tbl:housed_mapping}, the original 19 categories of labels were converted in batches to the 7 categories of this study.
For merged categories, a new target category label was uniformly assigned.
Professional annotators were organized to review the converted data, focusing on whether the bounding boxes were accurately fitted to the target edges and whether the merged category labels were semantically accurate.
Manual corrections were made for original HOUSED annotation errors or loose bounding boxes.
All final images were annotated with axis-aligned bounding boxes, recording: the issue category label (one of seven) and bounding box coordinates.

Addressing the severe long-tail distribution phenomenon in the integrated dataset, this study constructed a two-stage data balancing strategy based on the ablation experiment results in Section~\ref{Sec:balanced dataset} to cooperatively solve the issues of insufficient minority class samples and flooded majority class samples.
Simultaneously, this study proactively restricted the introduction of overly aggressive online strong data augmentations (such as MixUp, Copy-Paste), aiming to maintain the statistical stability of the semantic context of building damage.
Specifically: the first stage focuses on establishing a coarse-grained long-tail defense line, implementing targeted expansion using offline augmentation (including horizontal flipping, Gaussian noise injection, and brightness perturbation) for minority categories with scarce sample sizes;
the second stage fine-tunes according to the stratified refined sampling mechanism that made the most decisive contribution to global localization accuracy in the ablation experiments—implementing 100\% absolute retention for high-risk rare categories, 50\% probability retention for medium categories, and strict under-sampling cleaning with 10\% probability for flooded dominant categories (only retaining samples that do not co-occur with other categories), while compressing the retention ratio of background negative samples to 5\%.
The ultimately generated balanced dataset contains a total of 52,572 annotated instances.
During experimental deployment, this dataset was divided into a training set and a validation set according to a spatial heterogeneity ratio of approximately 8:2, serving as the benchmark data flow for subsequent multi-detector evaluation and post-correction fusion.
To quantitatively evaluate the processed data features, this study conducted a visual analysis of this training set from four dimensions (see Figure~\ref{fig3_total}).
Fig~\ref{fig3_1} shows that the processed category distribution has become reasonable, capable of supporting stable feature extraction by deep learning models. The spatial heatmaps in Fig~\ref{fig3_2} and Fig~\ref{fig3_3} reflect a concentration of issue targets in the center of the images.
The most critical feature is reflected in the size distribution in Fig~\ref{fig3_4}: the normalized width and height of the vast majority of targets are concentrated within the $[0.05, 0.20]$ range.

\section{Methods}
\subsection{Framework Overview}
Vision-based building inspection provides an important means of identifying housing-related issues and unauthorized modifications in old residential communities.
By analyzing field-acquired inspection images, modern object detection models can identify visible problems such as wall damage, pipeline deterioration, unauthorized renovation, and unauthorized balcony or window additions.
However, detection performance may be affected by complex backgrounds, occlusion, viewpoint variation, and incomplete image coverage.
More importantly, visual information alone cannot characterize the broader urban context surrounding a residential building.

A building's physical condition may be associated with both its internal characteristics and the functional environment of the surrounding community.
This study does not treat neighborhood context as a causal determinant of housing-related issues.
Instead, POI-derived urban context is used as a supplementary predictive prior to enhance the reliability of building-level inspection.

\begin{figure*}[b]
  \centering
  \includegraphics[width=0.9\textwidth]{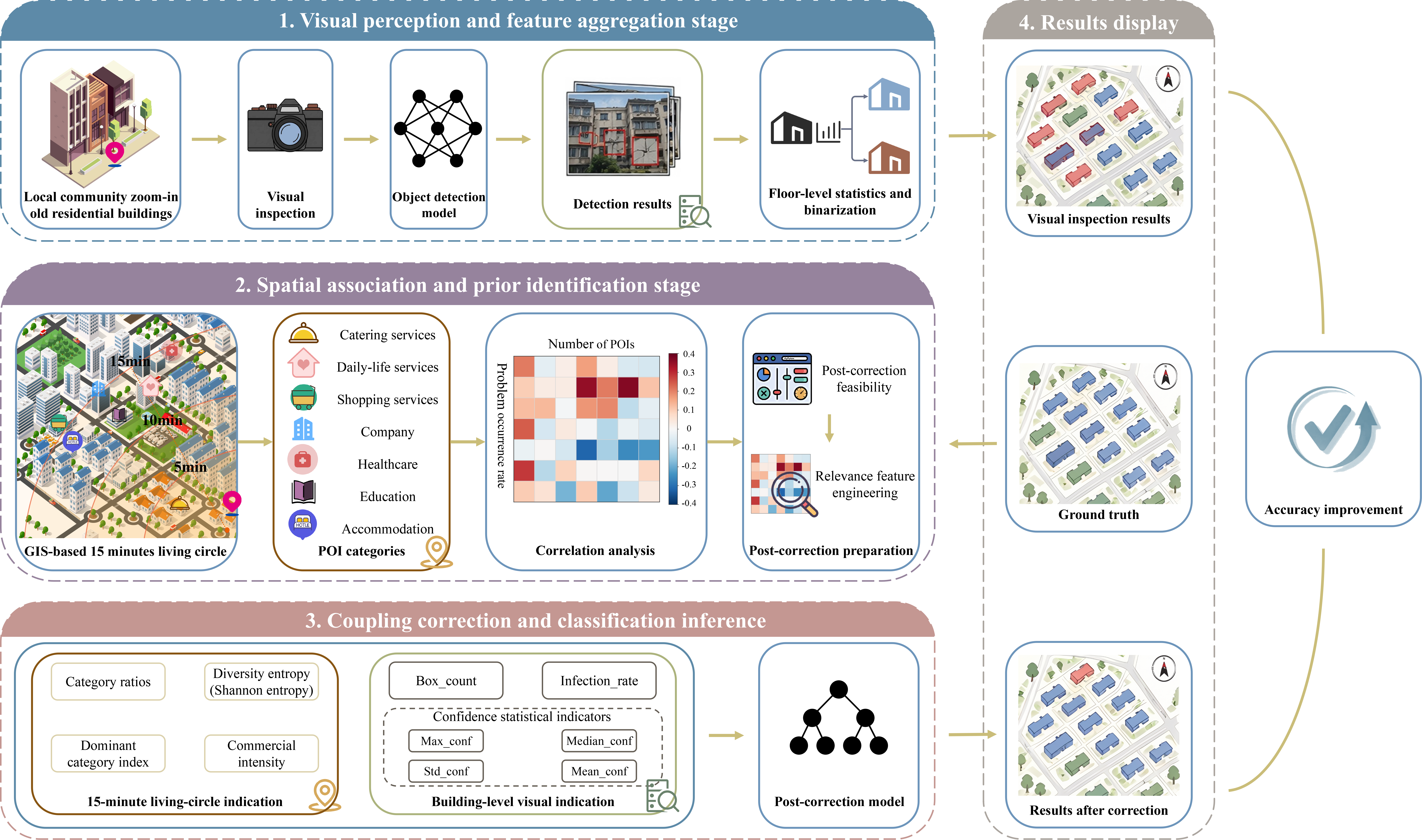} 
  \caption{Overall flow chart of the "perception-association-correction" three-stage vision--POI fusion framework for urban health check-ups.}
  \label{Method}
\end{figure*}

To support housing inspection and neighborhood-level interpretation in old residential communities, this study develops a three-stage framework consisting of visual perception, spatial association, and contextual post-correction.
The overall workflow is shown in Figure~\ref{Method}.

\begin{itemize} 
\item
Visual perception and feature aggregation stage:
First, the HOUSED dataset and the housing health check-up images of old communities in Qingdao collected in the field were integrated to construct a multi-source visual database;
subsequently, various object detection models were used to infer the images, outputting the position, category, and confidence scores of issue bounding boxes.
Since the decision-making objects of urban renewal work are buildings rather than single images, the detection results were aggregated at the building scale through an "image-level to building-level" spatial aggregation mechanism, generating a binary visual existence indicator and a 6-dimensional visual feature vector including detected object count, maximum confidence score, mean confidence score, median confidence score, standard deviation of confidence scores, and view-level occurrence rate, providing basic visual evidence for subsequent discrimination.
The output of this stage can also serve as an independent system prediction using only visual perception means, corresponding to the Direct Detection (DD) and Multi-View Visual Aggregation (MVVA) methods defined in the experiment section.
\item
Spatial association and prior identification stage:
To depict the community environment where residential buildings are located, the study used the 15-minute walking life circle as the maximum spatial boundary, constructed three POI search radii of 500~m, 1000~m, and 1500~m, and formed community-level POI feature matrices respectively.
Subsequently, the Pearson linear correlation coefficients and Spearman rank correlation coefficients between the proportions of the 7 POI categories and the incidence rates of the 7 health check-up issues were calculated at each spatial scale, and False Discovery Rate (FDR) correction was applied to control the risk of false positives in multiple testing.
This stage provides interpretable environmental prior information for subsequent models;
and the spatial radius finally entering the post-correction main model is further compared and determined by the multi-scale comparative experiments in Section~\ref{Sec:POI fusion}.
\item
Coupling correction and classification inference stage:
Finally, taking the building as the basic correction unit, an ensemble learning model with Random Forest as the classifier was constructed.
The building-level 6-dimensional visual features and the filtered community POI environmental features were cascaded and input, and hyperparameters were automatically optimized through grid search and cross-validation.
By introducing community environmental prior information, the proposed correction model is designed to investigate whether contextual information can help recover some visual false negatives and suppress some false positives, and suppress the false positives generated by image models, thereby enhancing the overall robustness and reliability of automated building condition detection.
\end{itemize}

Together, these three stages transform image-level visual observations into building-level inspection judgments.
visual perception provides direct physical evidence of buildings, spatial association extracts community environmental priors with statistical support,
and coupling correction fuses the two types of information to generate final building-level judgment results.
Compared with traditional detection workflows that rely solely on visual information, the method proposed in this paper can not only fully utilize the urban environmental background to improve detection accuracy, but also preserve the traceability of visual observation results and community environmental features, providing interpretable technical support for subsequent on-site verification, formulation of housing maintenance priorities, and urban renewal decision-making.

\subsection{Visual Perception and Feature Aggregation Stage}
The framework proposed in this paper completes visual perception at the image level and ultimately generates housing inspection results at the building level.
Since the object of housing condition assessment is residential buildings rather than single images, the image-level prediction results need to be converted into stable and reliable building-level representations before introducing environmental information for correction.
To achieve this goal, the visual perception module consists of two sequential components: Image-level Inspection Issue Detection and Building-level Feature Aggregation.
The former adopts deep object detection models to extract information on building inspection issues from a single inspection image,
whereas the latter fuses detection outputs from multiple perspectives to construct visual features that can characterize the overall condition of buildings, serving as input for subsequent classification.

\subsubsection{Image-level Visual Perception}
The visual perception stage employs object detection models to predict building facade images.
For the input image, the object detection model outputs the position, category, and corresponding confidence of each detected target as the foundational visual evidence for subsequent building-level inference.

Considering the significant differences in scale, appearance morphology, and surrounding environments of detection targets in old residential communities, this paper selected various state-of-the-art object detection models to systematically evaluate their applicability in housing condition detection tasks.
The models participating in the comparison include representative single-stage convolutional detectors YOLOv8, YOLOv9c, YOLOv10s, and YOLO11n, as well as the Transformer architecture-based detection model RT-DETR-L.
These models represent different object detection technical routes, thus enabling a relatively comprehensive evaluation of the robustness of the visual perception module under complex urban scenarios.
The specific training configurations of each model are detailed in Section~\ref{Sec:Experiment}.

Compared with traditional object detection datasets, the types of building health check-up issues possess particularity.
Some problem types (such as Elevator addition, Pipeline damage) have relatively stable visible physical structures; while other problem types (such as Wall damage, Public spaces illegal occupation, or No aging place modifications) are more discrete in visual manifestation and highly dependent on the surrounding scene context.
Therefore, a model achieving higher mAP at the image level may not necessarily provide the most informative building-level representation after building-scale aggregation.
Based on this, the detector evaluation in this study focuses not only on conventional image-level precision metrics but also more on the adaptability of these detection outputs in the subsequent building-level urban health check-up workflow.

\subsubsection{Building-level Feature Aggregation}
Since urban health check-ups target entire buildings rather than single images for governance, this study spatially aggregated image-level detection results to the building scale, enabling the post-correction model to possess bi-directional rectification capabilities of capturing visual false negatives and filtering visual false positives.

First, a binary Visual-existence indicator, $z$ , was introduced to indicate whether a building has direct visual evidence.
Specifically, $z=1$ if at least one detection of the target category is present in any inspection image associated with the building; otherwise, $z=0$.
For buildings with visual evidence ($z$=1), this paper extracts six statistical features from all bounding boxes belonging to the target category for that building.
These features describe the visual information of the building from multiple aspects, including detection count, prediction confidence, consistency, and spatial coverage degree, which are:
Detected Object Count;
Maximum Confidence Score;
Mean Confidence Score;
Median Confidence Score;
Standard Deviation of Confidence Scores;
View-level occurrence rate, the proportion of inspection images containing at least one target detection result to total images.

For buildings where no targets are detected ($z$=0), the above six statistical features are all assigned a value of 0, while retaining the visibility indicator $z$.
Therefore, each residential building can ultimately be represented as a unified building-level visual feature vector $x_v$ composed of one binary visibility indicator and six aggregated statistical features.
Compared with simple binary classification detection results, the aggregation strategy proposed in this paper can retain richer visual information.
Specifically, the detected object count reflects the overall degree of visible issues in the building;
multiple confidence statistical indicators depict the reliability of the model prediction results from different perspectives;
the view-level occurrence rate describes the consistency and spatial coverage degree of the issues appearing under different shooting angles.
The complete definitions of building-level aggregated features are shown in Table~\ref{tab:aggregated_features}.
\begin{table*}[htbp]
\centering
\caption{Definitions of the building-level visual features used in this study.}
\label{tab:aggregated_features}
\begin{tabularx}{\textwidth}{l X}
\toprule
\textbf{Feature Name} & \textbf{Definition} \\
\midrule
Detected Object Count                   & The total number of bounding boxes detected for the target category across all images of the building. \\
Maximum Confidence Score                & The highest confidence score among all detected bounding boxes for the target category. \\
Mean Confidence Score                   & The arithmetic mean of the confidence scores for all detected bounding boxes within the category. \\
Median Confidence Score                 & The median of the confidence scores for all detected bounding boxes within the category. \\
Standard Deviation of Confidence Scores & The standard deviation of the confidence scores for all detected bounding boxes within the category. \\
View-level occurrence rate                          & The ratio of the number of images with detected defects of this category to the total number of images for the building. \\
Visual Existence                        & A binary indicator: $z=1$ if detected object count $> 0$, and 0 otherwise. \\
\bottomrule
\end{tabularx}
\end{table*}

\subsection{Spatial Association and Environmental Prior Identification Stage}
\subsubsection{Community POI Life Circle Feature System}
This study adopts the '15-minute community life circle' as the maximum spatial boundary representing the functional context of the community.
The community center point is determined by the mean of the coordinates of all residential buildings within the community, and 500~m, 1000~m, and 1500~m buffer spaces are established respectively with this point as the center.
Among them, the 500~m buffer is used to depict direct life services and micro-commercial activities within the immediate neighborhood of the community;
the 1000~m buffer is used to represent a more complete block-level daily activity range;
the 1,500~m buffer approximates the upper spatial extent of the 15-minute neighborhood context considered in this study.
The design of multi-scale buffers aims to avoid pre-setting a single spatial scale as the optimal boundary, and to provide a consistent data basis for subsequently testing the impact of different POI context radii on the performance of the post-correction model.

This study designed a POI feature system composed of basic quantity features and high-order structural features (summarized as shown in Table~\ref{tab:poi_features}).
Among them, the basic feature is the total amount of POIs within the life circle.
To characterize the relative functional composition, this study converted high-dimensional POI raw records into relative structural indicators, and constructed a system containing the following four high-order features:

(1) Category Ratios: Converting the counts of the 7 basic POI categories into relative proportions to identify whether the life circle is dominated by commercial, life services, accommodation, healthcare, education, or corporate offices functions;

(2) Diversity entropy: Used to measure the mixing degree and balance of functions within the life circle, calculated based on the Shannon entropy formula (see Equation~\ref{E1_H}) to characterize the diversity and balance of neighborhood functions.;

(3) Dominant-category indicator: Marking the index of the POI category with the largest proportion, representing the core planning-dominant function of the community;

(4) Commercial intensity (CI): Defined as the sum of the proportions of Dining services and Shopping services (see Equation~\ref{E2_CI}), used to reflect the external load pressure brought by consumption-oriented activities to the community environment.
The above indicators jointly transform high-dimensional POI raw data into an interpretable community functional context portrait.
\begin{equation}
\label{E1_H}
    H = -\sum_{i=1}^{N} p_i \cdot \ln p_i
\end{equation}
\begin{equation}
\label{E2_CI}
    \mathit{CI} = p_{\text{dining}} + p_{\text{shopping}}
\end{equation}
Where Equation~\ref{E1_H} is the calculation formula for diversity entropy (Shannon entropy), $N$ represents the total number of POI categories examined within the life circle ($N$=7 in this study), and $p_i$ represents the normalized proportion of the $i$-th POI category among the 7 target POI categories ($\sum p_i = 1$).
Equation~\ref{E2_CI} is the calculation formula for CI, where $p_{dining}$ and $p_{shopping}$ respectively correspond to the proportions of Dining services and Shopping services.

\begin{table*}[htbp]
\centering
\caption{POI life circle feature system.}
\label{tab:poi_features}
\begin{tabularx}{\textwidth}{ll X}
\toprule
\textbf{Feature Level} & \textbf{Feature Name} & \textbf{Definition} \\
\midrule
Baseline Feature & Total POI count & Total number of POIs within the neighborhood pedestrian life circle. \\
\addlinespace 
Baseline Feature & POI count by category & Raw counts across 7 POI categories such as dining, lifestyle, and shopping. \\
\addlinespace
High-order Feature & POI ratio by category & Percentage of each POI category count relative to the total POI volume. \\
\addlinespace
High-order Feature & Diversity entropy & Shannon entropy. \\
High-order Feature & Dominant category indicator & Index of the POI category with the highest proportion. \\
High-order Feature & Commercial intensity & Sum of the proportions of dining and shopping services. \\
\bottomrule
\end{tabularx}
\end{table*}

\subsubsection{Multi-scale Correlation Analysis}
Correlation analysis aims to identify environmental variables that have significant statistical associations with the incidence rates of urban health check-up issues, thereby providing candidate urban context features for subsequent building-level post-correction.
Considering that there may be complex non-linear coupling relationships between community POI features and health check-up problem incidence rates, and some indicators do not strictly follow normal distributions, this study adopted a dual-testing strategy combining Pearson linear correlation and Spearman rank correlation.

(1) Pearson correlation analysis is used to capture linear associations between variables, calculated as shown in Equation~\ref{E3_Pearson}.
\begin{equation}
\label{E3_Pearson}
    r = \frac{\sum_{i=1}^{n} (x_i - \bar{x})(y_i - \bar{y})}{\sqrt{\sum_{i=1}^{n} (x_i - \bar{x})^2} \cdot \sqrt{\sum_{i=1}^{n} (y_i - \bar{y})^2}}
\end{equation}
Where $n$ is the total number of sample communities ($n$=92);
$x_i$ refers to the proportion of a specific POI category in the $i$-th community;
$y_i$ refers to the problem incidence rate under a specific health check-up problem in the $i$-th community;
$\bar{x}$ and $\bar{y}$ respectively correspond to the arithmetic means of the above two sets of variables in the 92 sample communities.
The correlation coefficient $r \in [-1,1]$, the larger its absolute value, the stronger the linear correlation, and the positive and negative signs represent positive and negative correlations, respectively.
This step is used to quantify the strength of linear association between POI features and problem incidence rates.

(2) Spearman rank correlation analysis does not require variables to follow a normal distribution, but captures monotonic but non-linear association patterns through rank transformation, thereby effectively enhancing the model's robustness to outliers, calculated as shown in Equation~\ref{E3_Spearman}.
\begin{equation}
\label{E3_Spearman}
    \rho = 1 - \frac{6 \sum_{i=1}^{n} d_i^2}{n(n^2 - 1)}
\end{equation}
Where $n$ is also the total number of sample communities ($n$=92); $d_i$ is the rank difference (i.e., ranking difference) of the two variables on the $i$-th community sample.
Unlike Pearson correlation coefficient which directly compares variable values, Spearman correlation coefficient measures the degree of correlation by comparing the consistency of variable rankings, and thus can identify monotonic associations that are difficult for linear models to depict.

This study constructed community-level statistical matrices at three spatial scales of 500m, 1000m, and 1500m respectively.
Completely identical statistical workflows were adopted at different scales (the complete correlation coefficient matrices are detailed in Figure 4-3 and Figure 4-4 in Section~\ref{Sec:POI fusion}).
To control the risk of false positives brought by multiple testing, the Benjamini-Hochberg method was adopted for FDR correction (screening threshold set to $p_{\text{FDR}} < 0.05$).
This step aims to identify candidate environmental variables with statistical significance, rather than directly determining the final POI radius; the best-performing scale for final modeling still needs to be verified through the building-level Multi-Source Context Fusion (MSCF) experiments in Section~\ref{Sec:POI fusion}.
The screening results of the three scales are shown in Table~\ref{tab:fdr_screening}.
\begin{table*}[htbp]
\centering
\caption{Summary of FDR screening results for three-scale correlation analysis.}
\label{tab:fdr_screening}
\begin{tabularx}{\textwidth}{
  l 
  >{\hsize=0.55\hsize\raggedright\arraybackslash}X 
  >{\hsize=0.75\hsize\raggedright\arraybackslash}X 
  >{\hsize=1.25\hsize\raggedright\arraybackslash}X 
  >{\hsize=1.25\hsize\raggedright\arraybackslash}X
}
\toprule
\textbf{Radius} & \textbf{Number of significant Pearson pairs} & \textbf{Number of significant Spearman pairs} & \textbf{Statistical interpretation} & \textbf{Subsequent processing} \\
\midrule
500m & 0 & 0 & Computed fully, but did not pass FDR screening & Served as a neighborhood-scale control; not listed in the significant pairs breakdown \\
\addlinespace
1000m & 0 & 2 & A small number of stable rank correlations appeared & Served as block-level candidate context; continued into modeling comparison \\
\addlinespace
1500m & 4 & 11 & Most significant pairs, stronger interpretability & Retained significant variables; the final radius is still determined by building-level F1 \\
\bottomrule
\end{tabularx}
\end{table*}
\begin{table*}[htbp]
\centering
\caption{Significant POI-health check-up problem pair groups after FDR correction.}
\label{tab:significant_poi_pairs}
\begin{tabularx}{\textwidth}{
  >{\hsize=1.0\hsize\raggedright\arraybackslash}X 
  >{\hsize=0.6\hsize\raggedright\arraybackslash}X 
  >{\hsize=0.6\hsize\raggedright\arraybackslash}X 
  >{\hsize=1.8\hsize\raggedright\arraybackslash}X
}
\toprule
\textbf{Problem Type} & \textbf{Radius} & \textbf{Method} & \textbf{\begin{tabular}[c]{@{}c@{}}Significant POI variables \\ (coefficients)\end{tabular}} \\
\midrule
Addition balconies windows 
& 1000m & Spearman 
& Education/Culture: $\rho=+0.323$; \newline Shopping: $\rho=-0.323$ \\
\addlinespace
& 1500m & Pearson; \newline Spearman 
& Accommodation: $r=+0.326$, $\rho=+0.335$; \newline Daily Service: $r=-0.338$, $\rho=-0.310$; \newline Shopping: $\rho=-0.313$; Dining: $\rho=-0.338$ \\
\addlinespace
Public spaces illegal occupation 
& 1500m & Spearman 
& Healthcare: $\rho=+0.296$; \newline Education/Culture: $\rho=+0.340$ \\
\addlinespace
Wall damage 
& 1500m & Spearman 
& Corporate offices: $\rho=+0.292$; \newline Healthcare: $\rho=+0.267$ \\
\addlinespace
Illegal renovation expansion 
& 1500m & Spearman 
& Healthcare: $\rho=+0.264$ \\
\addlinespace
Elevator addition 
& 1500m & Pearson; \newline Spearman 
& Shopping: $r=+0.431$, $\rho=+0.324$; \newline Education/Culture: $r=-0.298$, $\rho=-0.368$ \\
\bottomrule
\end{tabularx}
\end{table*}
From Table~\ref{tab:fdr_screening}, it can be seen that at the 500m radius, neither Pearson nor Spearman tests yielded significant correlated pairs, indicating that the POI proportions at the neighborhood scale are insufficient to form stable statistical associations under the current community sample size;
at the 1000m radius, only 2 significant rank correlated pairs appeared in the Spearman test, suggesting that the block-scale context begins to provide supplementary explanation;
at the 1500m radius, the number of significant pairs is the most, indicating that a larger life circle range is more conducive to capturing statistical patterns between community functional structures and problem incidence rates.

Table~\ref{tab:significant_poi_pairs} lists the Pearson correlation coefficients r or Spearman correlation coefficients rho that are significant ($p_{\text{FDR}} < 0.05$) and have large absolute values after FDR correction.
Further dissecting the spatial socioeconomic mechanisms of the significant associations in Table~\ref{tab:significant_poi_pairs}, it can be found that:
first, Addition balconies windows are significantly positively correlated with Education/Culture and Accommodation at the 1000m and 1500m scales, which is primarily driven by the rental interests of highly mobile populations such as shared renting around universities and short-term renting in scenic spots;

whereas their significant negative correlation with Dining and Daily Service is because highly commercialized blocks face denser urban management grid inspections and facade control constraints.
Second, Public spaces illegal occupation is significantly positively correlated with Healthcare and Education/Culture, reflecting the long-standing tidal convergence of people and vehicles around hospitals and schools, and the imbalance between supply and demand of parking resources, which easily triggers the disorder of public spaces.

It is particularly important to note that although the 1500m spatial radius screened out the most significant correlated pairs under FDR correction, this does not mean it is the best-performing scale for micro-prediction.
The statistical tests in this section focus on proving the effectiveness of macroscopic environmental priors;
whereas in the multivariate building-level prediction conducted in Section~\ref{Sec:POI fusion}, the 1500m scale is highly prone to crossing main roads and introducing heterogeneous functional noise from adjacent blocks, weakening the spatial specificity of specific buildings.
Therefore, the strongest macroscopic correlation is not equivalent to the best micro-prediction performance. The best-performing feature fusion scale in this dataset will be determined by the ensemble learning model in Section~\ref{Sec:POI fusion}.

\subsection{MSCF and Post-Correction Mechanism}
Based on the extracted visual aggregated features and environmental prior features, this study constructed the MSCF post-correction framework. Let the 6-dimensional visual feature vector of a building be $x_v$ (i.e., the building-level aggregated features defined in Section 3.2.2), and the POI environmental feature vector of its affiliated community at a specific scale be $x_e$ (the feature set screened based on the correlation analysis in Section 3.3.2), the post-correction judgment process can be formalized as (see Equation~\ref{P}):
\begin{equation}
\label{P}
P(y_b = 1 \mid x_v, x_e) = f(x_v, x_e)
\end{equation}
Where $y_b \in \{0, 1\}$ expresses the judgment state of the building's specific health check-up issue ($y_b$=1 indicates the presence of inspection issues, 0 indicates normal); 
$P(y_b = 1 \mid x_v, x_e)$ is the post-correction conditional probability;
$f(\cdot)$ corresponds to the feature mapping and probability prediction function of the post-correction classifier.
The final building judgment state $y_b$ is determined by partitioning this conditional probability through a decision threshold $\theta$, i.e., when the probability value is greater than or equal to $\theta$, it is judged that a inspection issue exists ($y_b$=1), otherwise it is judged as normal ($y_b$=0).

In concrete implementation, the post-correction mechanism employs Random Forest to independently train a binary classifier for each of the 7 core health check-up problem categories.
The model takes the cascaded visual features and environmental feature vector [$x_v$, $x_e$] as input, utilizes grid search and cross-validation to automatically optimize hyperparameters, and introduces a Cost-Sensitive Learning mechanism by setting compensatory class weights to offset the imbalance of sample distribution.
The specific hyperparameter search space for the Random Forest classifier is detailed in Table \ref{tab:hyperparameter_space}. Grid search adopts 3-fold cross-validation, with the evaluation metric being F1-Score. The final model adopts 5-fold cross-validation for performance evaluation to ensure the robustness of the results.
\begin{table}[htbp]
\centering
\caption{Hyperparameter search space of the post-correction classifiers}
\label{tab:hyperparameter_space}
\setlength{\tabcolsep}{4pt} 
\small
\begin{tabularx}{\linewidth}{
  @{}l
  >{\raggedright\arraybackslash\hsize=1.05\hsize}X
  >{\raggedright\arraybackslash\hsize=0.95\hsize}X@{}
}
\toprule
\textbf{Hyperparameter} & \textbf{Search Space} & \textbf{Description} \\
\midrule
n\_estimators & [100, 150, 200, 250, 300, 350, 400] & Number of decision trees \\
\addlinespace
max\_depth & [4, 6, 8, 12, 16, 20] & Maximum depth of the tree \\
\addlinespace
min\_samples\_split & [2, 5, 8, 10, 12] & Minimum number of samples required to split an internal node \\
\addlinespace
min\_samples\_leaf & [1, 2, 4, 6, 8] & Minimum number of samples at a leaf node \\
\bottomrule
\end{tabularx}
\end{table}
\subsection{Method Output and Evaluation Metrics}
\subsubsection{Multi-level Method Output}
The method output framework constructed in this study covers three spatial levels from micro to macro, and after connection, transforms the visual recognition process into decision-making information usable for urban health check-ups.
The specific output levels include:

(1) Image-level detection output: Records the position (Bounding Box coordinates), category, and detection confidence of problem targets in single facade images, providing micro evidence for visual verification and manual review.

(2) Building-level statistical output: Aggregates the detection results of all associated images of the same building to form a 6-dimensional feature vector (e.g., detected object count, max/mean confidence, etc.) and a binary judgment state of issue presence, providing an inspection inventory for urban health check-ups taking the building as the basic unit.

(3) Community-level coupling output: Correlates building-level disease data with community-scale POI environmental features, revealing the environmental associations between the spatial distribution of housing-related issues and the community functional structure, providing spatial decision-making basis for regional-scale urban governance and renewal planning.
\subsubsection{Evaluation Metrics}
To comprehensively verify the effectiveness of the perception, association, and correction stages, this study constructed an evaluation metric system covering the three-dimensional spatial levels of "image-building-community" (the correspondence of metrics for each evaluation method is detailed in Table~\ref{tab:evaluation_metrics}):

(1) Image-level object detection evaluation: Precision, Recall, F1-score and mean Average Precision (mAP) are adopted to evaluate the detection model for building inspection issues.
For image-level Precision, Recall, and F1-score, an Intersection over Union (IoU) threshold greater than 0.15 was used as a coarse evidence-matching criterion between predicted and annotated issue regions.
This threshold was selected for facade inspection targets because such issues often have irregular shapes, partial visibility, perspective distortion, and ambiguous annotation boundaries.
It was used to assess whether a predicted box provided visual evidence for a target issue, rather than to replace strict localization evaluation.
Strict localization capability was therefore reported separately using mAP50 (IoU=0.5) and mAP50-95 (IoU averaged from 0.5 to 0.95 with a step of 0.05).

(2) Environmental association efficacy evaluation: Utilizing Pearson linear correlation coefficient r and Spearman rank correlation coefficient rho to quantify the statistical associations between the proportion features of each POI category and the incidence rates of 7 health check-up issues under 500m, 1000m, and 1500m radii, and combining Benjamini-Hochberg FDR correction to screen out significant spatial context features.

(3) Building-level post-correction decision evaluation: At the building level, the post-correction model combines image-level aggregated features and POI environmental features for judgment. The evaluation adopted Precision, Recall, Accuracy, and F1-score as efficacy indicators for decision binary judgment;
meanwhile, the Area Under the ROC Curve (ROC-AUC) and Area Under the PR Curve (PR-AUC) were introduced to comprehensively quantify the probabilistic discriminability outputted by the post-correction model, serving to comparatively evaluate the prediction performance of four post-correction strategies on the test set: Direct Detection (DD), Multi-View Visual Aggregation (MVVA), Urban Functional Context Only (UFCO), and Multi-Source Context Fusion (MSCF).
These comparisons quantify the rectification effectiveness of multi-source information fusion on visual missed detections and false alarms.
\begin{table*}[htbp]
\centering
\caption{Evaluation metrics.}
\label{tab:evaluation_metrics}
\begin{tabularx}{\textwidth}{
  >{\hsize=0.6\hsize\raggedright\arraybackslash}X 
  l 
  l
  >{\hsize=1.0\hsize\raggedright\arraybackslash}X 
  >{\hsize=1.4\hsize\raggedright\arraybackslash}X
}
\toprule
\textbf{Evaluation Dimension} & \textbf{Metric} & \textbf{Abbreviation} & \textbf{Calculation Method} & \textbf{Evaluation Purpose} \\
\midrule
Image Recognition 
& mAP50 & mAP@0.5 & Mean average precision at IoU=0.5 & Comprehensive accuracy of localization and recognition \\
& mAP50-95 & mAP@0.5:0.95 & Mean average precision at IoU=0.5$\sim$0.95 & High-precision boundary box localization capability \\
& Precision & $P$ & $TP / (TP + FP)$ & Precision of bounding box predictions \\
& Recall & $R$ & $TP / (TP + FN)$ & Model's ability to identify true issue instances \\
\addlinespace
Environmental Association 
& \begin{tabular}[c]{@{}l@{}}Pearson $r$ / \\ Spearman $\rho$\end{tabular} & $r / \rho$ & Linear/Rank correlation coefficient & Strength of statistical association \\
\addlinespace
Building Judgment 
& Precision & $P$ & $TP / (TP + FP)$ & Precision of post-correction binary judgments \\
& Recall & $R$ & $TP / (TP + FN)$ & Recall capability for true issues after correction \\
& F1-score & $F1$ & $2 P R / (P + R)$ & Comprehensively balanced effect of post-correction judgments \\
& Accuracy & $Acc$ & $(TP + TN) / N$ & Overall accuracy of post-correction binary judgments \\
& ROC-AUC & $AUC_{ROC}$ & Area under the ROC curve & Discrimination capability of probability outputs (threshold-independent) \\
& PR-AUC (AP) & $AUC_{PR}$ & Area under the PR curve & Classification performance under severe class imbalance \\
\addlinespace
Efficiency Evaluation 
& Throughput & $FPS$ & Images processed per second (samples/s) & Actual speed of forward inference \\
& Peak VRAM & $VRAM$ & Peak GPU memory usage (GB) & Peak spatial complexity during model inference \\
\bottomrule
\end{tabularx}
\end{table*}

\section{Experiments and Results}
In order to systematically verify the effectiveness and robustness of the proposed method, this section designs experiments around the three-level progressive logic of "visual perception — cross-scale aggregation — multi-modal contextual correction".
Since the ultimate goal of urban health check-ups is to provide decision support for refined governance, this chapter elevates the evaluation horizon from local pixels of single images to physical building units that are spatially locatable and mechanistically interpretable.

The experimental content of this study is mainly divided into the following three levels:
(1) Visual perception benchmark and ablation (image-level): Comparing the perception efficacy of mainstream object detection algorithms on complex health check-up datasets, and evaluating the impact of data balancing and augmentation strategies on model robustness.
(2) Cross-scale aggregation analysis (building-level): Based on the probabilistic soft aggregation mechanism, the image detection results are mapped to physical building units to explore the characterization ability of multi-view aggregated features for overall building inspection issues.
(3) Contextual post-correction verification (community and building-level): Integrating multi-scale community POI statistics to examine how the ensemble learning framework effectively corrects the classification bias of front-end visual models.
To ensure the rigor of the comparison, all benchmark models were trained under unified data partitioning and hyperparameter protocols to eliminate confounding variable interference.
In the model comparison, the visual network with a 640-pixel input resolution was used as the backbone benchmark, supplemented by a high-resolution experiment of 1280 pixels, to explore the potential marginal benefits of image scale while considering computational feasibility.

For readability, Table~\ref{tab:key_abbreviations} summarizes the main abbreviations used in the experimental framework.
These terms distinguish image-only baselines, context-only tests, fused decision models, and validation metrics.

\begin{table}[htbp]
\centering
\caption{Key abbreviations used in the experimental framework.}
\label{tab:key_abbreviations}
\begin{tabularx}{\linewidth}{@{} l X @{}}
\toprule
\textbf{Abbreviation} & \textbf{Meaning and role in this study} \\
\midrule
POI & Point of Interest; used to describe the surrounding urban functional context. \\
DD & Direct Detection; a building is judged positive if any associated image contains a detected target box. \\
MVVA & Multi-View Visual Aggregation; aggregates detection evidence from all images of the same building into visual statistical features. \\
UFCO & Urban Functional Context Only; uses only POI-derived context features without visual evidence. \\
MSCF & Multi-Source Context Fusion; combines MVVA visual features with POI context features for building-level diagnosis. \\
Spatial Group CV & Spatial Group Cross-Validation; keeps buildings from the same community in the same fold to reduce spatial leakage. \\
FDR & False Discovery Rate; used to control multiple-testing error in POI correlation screening. \\
PR-AUC & Precision-Recall Area Under the Curve; evaluates probability ranking under class imbalance. \\
\bottomrule
\end{tabularx}
\end{table}

\subsection{Experimental Environment}
\label{Sec:Experiment}
To ensure the repeatability of experiments and the rigor of performance evaluation, all deep learning training and inference tasks in this study were executed on a high-performance computing workstation equipped with four NVIDIA GeForce RTX 3090 GPUs (24 GB memory each).
The software environment was built on Python 3.10.19, PyTorch 2.5.1, and CUDA 12.1.
The visual object detection models were implemented based on the Ultralytics 8.4.6 framework, while the post-correction ensemble models (Random Forest) were built using scikit-learn with grid search cross-validation.
Under this benchmark hardware architecture, this study constructed an end-to-end computational throughput and resource overhead monitoring protocol.
Utilizing AcademicTracker, we dynamically tracked and recorded the algorithm's throughput, Peak GPU Memory, Parameter Count, and floating-point operations (FLOPs) during the training and inference stages, thereby providing objective computational efficacy support for the practical deployment feasibility of subsequent multi-models in old community health check-up tasks.

\subsection{Ablation of Data Balancing and Augmentation Strategies}
\label{Sec:balanced dataset}
Before conducting cross-architecture benchmark evaluations of detectors, this study first takes YOLOv8 as the ablation baseline to systematically explore the synergistic effects of data distribution balancing and spatial synthetic augmentation strategies on the detection of multi-dimensional building inspection issues in housing health inspection scenarios.
The data subset for ablation experiments strictly follows the balanced sample space established in Section 2.4, containing a total of 52,572 independent disease annotation instances. Its partitioning structure is detailed in Table~\ref{tab:dataset_scale}.
\begin{table}[htbp]
\centering
\caption{Number of images and scale of annotated instances in the balanced dataset.}
\label{tab:dataset_scale}
\begin{tabularx}{\linewidth}{
  >{\raggedright\arraybackslash}X 
  >{\centering\arraybackslash}X 
  >{\centering\arraybackslash}X
}
\toprule
\textbf{Split} & \textbf{Images} & \textbf{Instances} \\
\midrule
Training & 11630 & 41964 \\
\addlinespace
Validation & 2892 & 10608 \\
\bottomrule
\end{tabularx}
\end{table}

This paper adopts a two-factor experimental design to systematically evaluate the impact of data distribution and augmentation strategies.
The first factor is data distribution, including the original long-tail distribution dataset and the balanced resampling dataset;
the second factor is the data augmentation strategy, including:
(i) traditional single-image augmentation method;
(ii) strong augmentation composite strategy.
Among them, traditional augmentation refers to image-level transformation operations performed online during training, including random horizontal flipping, scaling, brightness adjustment, and contrast perturbation, etc., used to simulate changes in imaging conditions.
In contrast, the strong augmentation strategy introduces cross-sample and instance-level synthesis methods, including MixUp and Copy-Paste, to generate composite training samples through linear mixing or object-level pasting, thereby significantly increasing data diversity.
Table~\ref{tab:ablation_results} summarizes the results of the four experimental configurations (A-1 to A-4).
To ensure fairness, all experiments adopted the same training settings, including batch size (16), optimizer configuration, and learning rate scheduling strategy.
\begin{table*}[htbp]
\centering
\caption{YOLOv8 data strategy ablation design and validation results.}
\label{tab:ablation_results}
\begin{tabular}{ccccccc}
\toprule
\textbf{\begin{tabular}[c]{@{}c@{}}Scheme \\ ID\end{tabular}} & 
\textbf{\begin{tabular}[c]{@{}c@{}}Training Dataset \\ Distribution Protocol\end{tabular}} & 
\textbf{\begin{tabular}[c]{@{}c@{}}Data Augmentation \\ Mechanism\end{tabular}} & 
\textbf{\begin{tabular}[c]{@{}c@{}}mAP@0.5 \\ (\%)\end{tabular}} & 
\textbf{\begin{tabular}[c]{@{}c@{}}mAP@0.5:0.95 \\ (\%)\end{tabular}} & 
\textbf{\begin{tabular}[c]{@{}c@{}}$\Delta$mAP@0.5 \\ (\%)\end{tabular}} & 
\textbf{\begin{tabular}[c]{@{}c@{}}$\Delta$mAP@0.5:0.95 \\ (\%)\end{tabular}} \\
\midrule
\begin{tabular}[c]{@{}c@{}}A-1\\ (Baseline)\end{tabular} 
& \begin{tabular}[c]{@{}c@{}}Original \\ Long-tailed Space\end{tabular} & (i) & 82.05 & 59.94 & -- & -- \\
\addlinespace
A-2 & \begin{tabular}[c]{@{}c@{}}Original \\ Long-tailed Space\end{tabular} & (ii) & 84.09 & 63.79 & +2.04 & +3.85 \\
\addlinespace
\textbf{\begin{tabular}[c]{@{}c@{}}A-3\\ (Optimal)\end{tabular}} 
& \textbf{\begin{tabular}[c]{@{}c@{}}Balanced \\ Resampling Space\end{tabular}} & \textbf{(i)} & \textbf{84.46} & \textbf{68.05} & \textbf{+2.41} & \textbf{+8.11} \\
\addlinespace
A-4 & \begin{tabular}[c]{@{}c@{}}Balanced \\ Resampling Space\end{tabular} & (ii) & 84.30 & 66.81 & +2.25 & +6.87 \\
\bottomrule
\end{tabular}
\end{table*}
\begin{figure}
	\centering
		\includegraphics[scale=0.55]{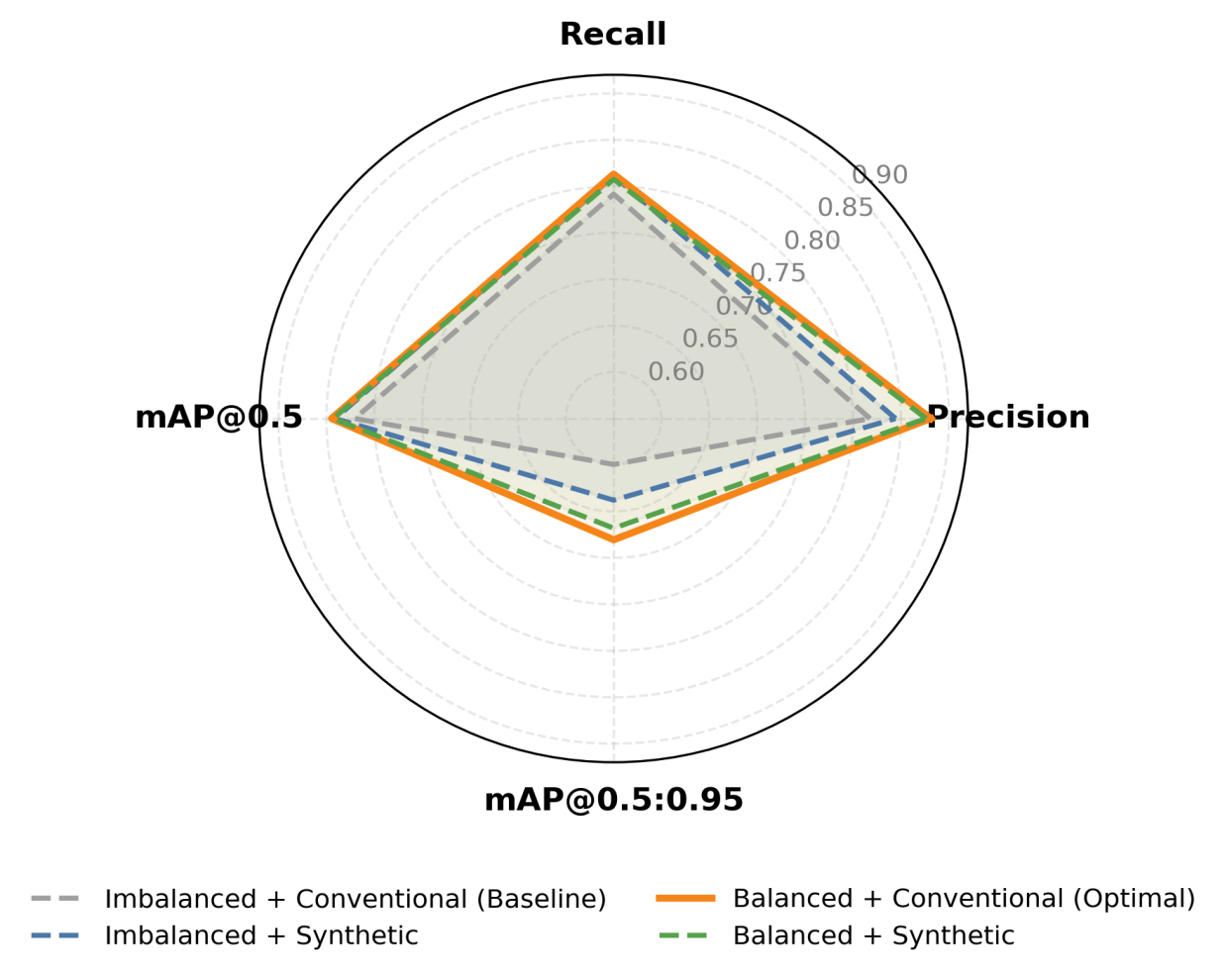}
	\caption{Radar chart of YOLOv8 validation performance under different data balancing and augmentation strategies.}
	\label{fig4_1_radar}
\end{figure}

The quantitative data analysis in Table~\ref{tab:ablation_results}, coupled with the significant area expansion of the radar chart in Figure~\ref{fig4_1_radar}, collectively demonstrate that balancing the data distribution significantly improves detection performance.
Specifically, after adjusting the original long-tail distribution (A-1) to the balanced resampling strategy (A-3), mAP@0.5 increased from 82.05\% to 84.46\%, and mAP@0.5:0.95 saw a substantial increase from 59.94\% to 68.05\%.
The obvious expansion of the radar chart area further intuitively verifies that such data distribution balancing can not only effectively mitigate recall bias and missed detections of minority inspection issue categories induced by long-tailed distributions, but also fundamentally enhance the detector’s robustness in edge localization of inspection issues.
This enhanced robustness is clearly corroborated by the more pronounced improvement under the stricter evaluation metric (mAP@0.5:0.95).
Furthermore, under the original long-tail distribution conditions, the strong augmentation strategy (A-2) increased mAP@0.5:0.95 to 63.79\%, indicating that under conditions of sample scarcity, composite augmentation played an effective regularization role by increasing data diversity.
However, under the balanced data distribution condition, the strong augmentation strategy did not yield additional benefits (A-4 compared to A-3), but instead slightly reduced performance (mAP@0.5:0.95 dropping to 66.81\%).
This suggests that when the training data is already relatively balanced, excessively strong synthetic perturbations may introduce distribution inconsistencies, thereby affecting the model's feature learning efficacy.

The experimental results indicate that there is an obvious interaction between data distribution and augmentation strategies.
Strong augmentation has a clear advantage under long-tail distribution conditions, but its effect weakens after data balancing is completed.
This indicates that the effectiveness of instance-level synthetic augmentation depends on the underlying statistical structure of the dataset.
From a methodological perspective, balanced resampling has already improved the category coverage problem to a certain extent, so the model's reliance on strong data augmentation is reduced.
In this case, adopting a simple augmentation strategy based on traditional image-level transformations can provide a more stable training process.
Based on the above results, subsequent experiments in this study all adopted balanced resampling combined with basic single-image real-time dynamic augmentation as a unified data processing protocol.

\subsection{Image-level Detector Performance}
\label{sec4.3}
\begin{figure}
	\centering
		\includegraphics[scale=0.35]{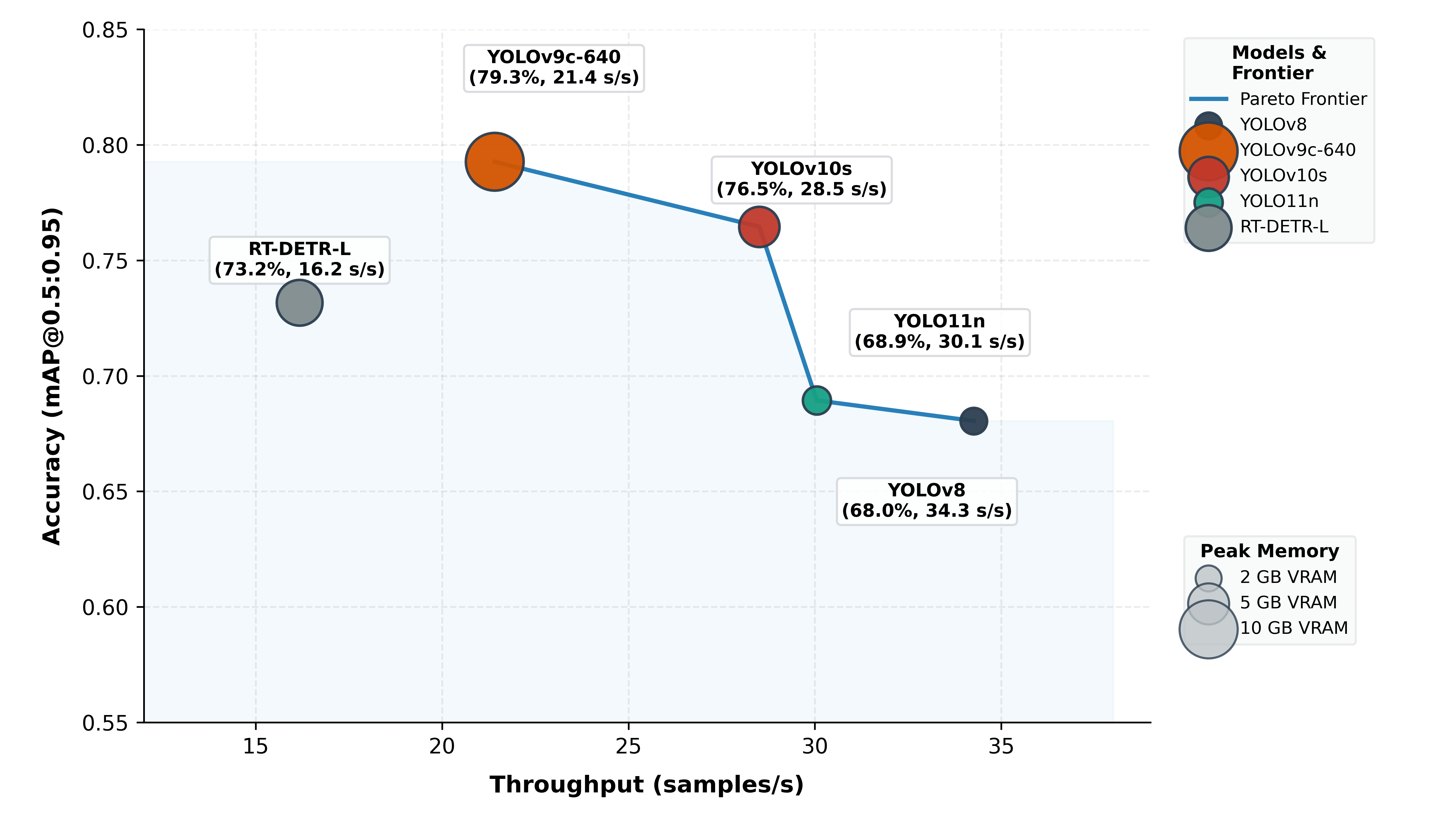}
	\caption{Comparison of image-level mAP of mainstream visual detectors.}
	\label{fig:4-2-mAP}
\end{figure}

Based on following the unified data resampling and augmentation strategy described in Section~\ref{Sec:balanced dataset}, this section evaluates the performance of different visual detection architectures at the image level.
Its goal is to examine the representation capabilities of each detector under a purely visual setting, providing a foundational basis for subsequent building-level feature aggregation and spatial context correction, rather than serving as the final decision model.
This paper constructed a comprehensive benchmark test containing five representative object detection models, including YOLOv8, YOLOv9c-640, YOLOv10s, YOLO11n, and RT-DETR-L.
These models cover CNN-based single-stage detectors and the Transformer-based detection paradigm, enabling a systematic comparison of different architectural designs in complex urban inspection scenarios.
The related detection precision and computational efficiency results are shown in Table~\ref{tab:performance_efficiency} and Figure~\ref{fig:4-2-mAP}, respectively.
\begin{table*}[htbp]
\centering
\caption{Overall performance and computational efficiency of mainstream visual detectors on the validation set}
\label{tab:performance_efficiency}
\begin{tabular}{ccccccc}
\toprule
\multirow{2}{*}{\textbf{Model}} & 
\multicolumn{4}{c}{\textbf{Accuracy Metrics}} & 
\multicolumn{2}{c}{\textbf{Computational Complexity}} \\
\cmidrule(lr){2-5} \cmidrule(lr){6-7}
& \textbf{P} & \textbf{Re} & \textbf{mAP@0.5} & \textbf{mAP@0.5:0.95} & 
\textbf{\begin{tabular}[c]{@{}c@{}}Throughput \\ (samples/s)\end{tabular}} & 
\textbf{\begin{tabular}[c]{@{}c@{}}Peak \\ VRAM \\ (GB)\end{tabular}} \\
\midrule
YOLOv8      & 88.11\% & 81.31\% & 84.46\% & 68.05\% & 34.26 & 2.09 \\
\addlinespace
YOLOv9c-640 & 92.14\% & 83.00\% & 88.11\% & 79.27\% & 21.41 & 9.88 \\
\addlinespace
YOLOv10s    & 92.67\% & 81.91\% & 86.85\% & 76.46\% & 28.51 & 4.82 \\
\addlinespace
YOLO11n     & 88.52\% & 81.73\% & 85.07\% & 68.94\% & 30.05 & 2.34 \\
\addlinespace
RT-DETR-L   & 92.43\% & 82.74\% & 86.47\% & 73.17\% & 16.18 & 6.20 \\
\bottomrule
\end{tabular}
\end{table*}

As shown in Table~\ref{tab:performance_efficiency}, YOLOv9c-640 achieved the best overall detection precision, reaching 88.11\% on mAP@0.5 and 79.27\% on mAP@0.5:0.95.
YOLOv10s and RT-DETR-L followed closely behind. RT-DETR-L demonstrated relatively balanced precision-recall performance.
However, its performance under stricter IoU thresholds was slightly lower than CNN detections.
In contrast, although YOLOv8 and YOLO11n had relatively lower overall precision, they exhibited more stable performance with smaller variance across different categories.

To further evaluate the impact of input resolution on detection performance, this paper conducted supplementary experiments on YOLOv9c-640 at a higher resolution of 1280×1280, with results shown in Table~\ref{tab:resolution_comparison}.
The results showed that increasing the input resolution only brought very limited performance improvements, with mAP@0.5 increasing by 0.13\% and mAP@0.5:0.95 increasing by 0.05\%.
At the same time, the recall rate slightly decreased, indicating limited overall improvement in detection robustness.
This result shows that in housing condition inspection tasks, increasing image resolution does not significantly enhance model performance.
For structural is such as illegal balcony additions and facade renovations, a resolution of 640×640 is sufficient to provide adequate geometric information. 
Higher resolutions are mainly somewhat helpful for issues sensitive to micro-textures, but in large-scale urban inspection tasks, their benefits show obvious marginal diminishing.
Therefore, balancing computational efficiency and accuracy gains, this paper uniformly adopted 640×640 as the default input resolution in all subsequent experiments.
\begin{table*}[htbp]
\centering
\caption{Performance comparison of YOLOv9c under different input resolutions}
\label{tab:resolution_comparison}
\begin{tabular}{lccccc}
\toprule
\textbf{Model} & \textbf{Input Size} & \textbf{P} & \textbf{R} & \textbf{mAP@0.5} & \textbf{mAP@0.5:0.95} \\
\midrule
\textbf{YOLOv9c} & \textbf{640 px} & \textbf{92.14\%} & \textbf{83.00\%} & \textbf{88.11\%} & \textbf{79.27\%} \\
\addlinespace
YOLOv9c & 1280 px & 92.93\% & 82.80\% & 88.24\% & 79.32\% \\
\addlinespace
Diff (1280 px vs 640 px) & -- & +0.79\% & -0.20\% & +0.13\% & +0.05\% \\
\bottomrule
\end{tabular}
\end{table*}

It is worth emphasizing that image-level detection performance cannot be directly equated with physical building-level health check-up judgment performance.
Image-level mAP mainly evaluates the comprehensive accuracy of bounding box spatial localization and category prediction, while the building-level post-correction module relies more on multi-dimensional spatial aggregated features composed of bounding box counts, confidence statistics, and view-level occurrence rates.
A visual detector's high mAP performance at the image level does not necessarily provide the most robust and discriminative aggregated feature distribution for the building-level post-correction classifier.
Therefore, in the evaluation system, this study positioned image-level mAP as a benchmark metric measuring "perceptual quality evidence", and positioned building-level F1-score and spatial cross-validation results as empirical metrics characterizing "urban governance decision efficacy".
The two types of metrics complement each other to jointly constitute a systematic empirical evaluation system.

\subsection{Multi-Scale POI Fusion Post-Correction Experiments}
\label{Sec:POI fusion}

\begin{figure*}[t!]
\centering
  \begin{subfigure}{0.48\textwidth}
    \centering
    \includegraphics[width=\linewidth]{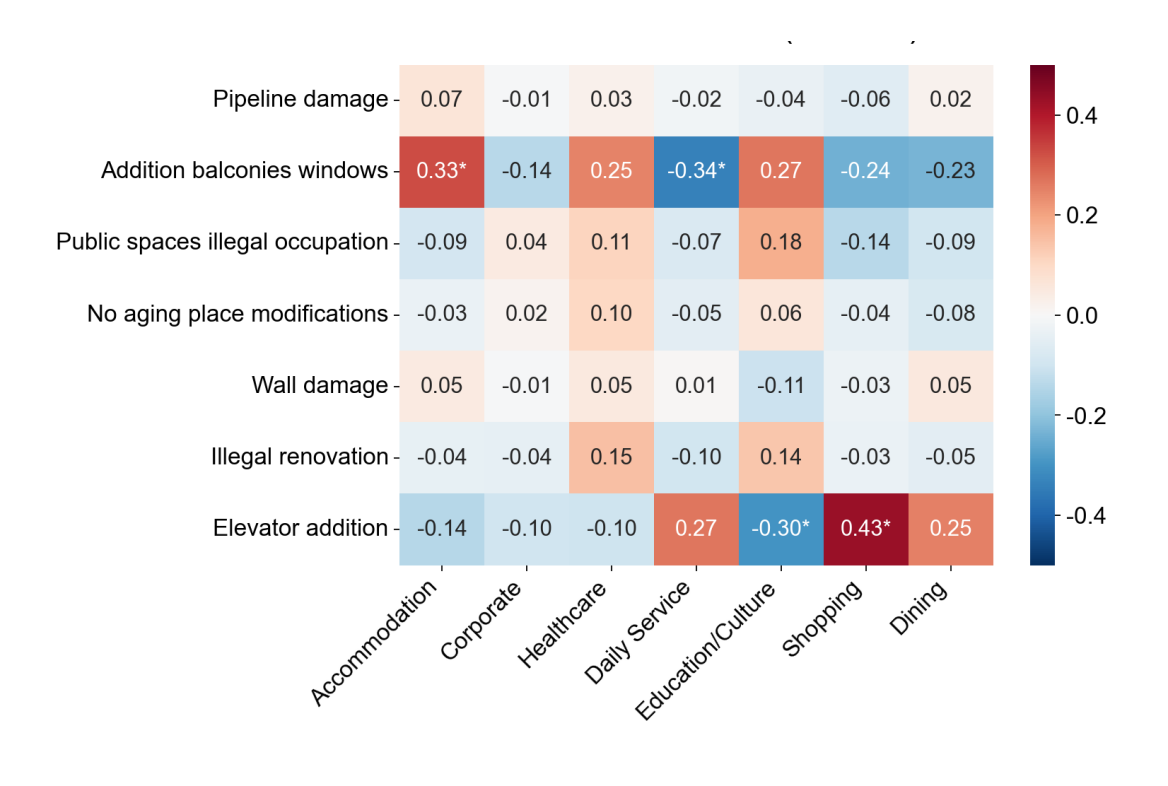}
    \caption{Pearson linear correlation (1500m)}
    \label{fig:sub_pearson_1500}
  \end{subfigure}
  \hfill
  \begin{subfigure}{0.48\textwidth}
    \centering
    \includegraphics[width=\linewidth]{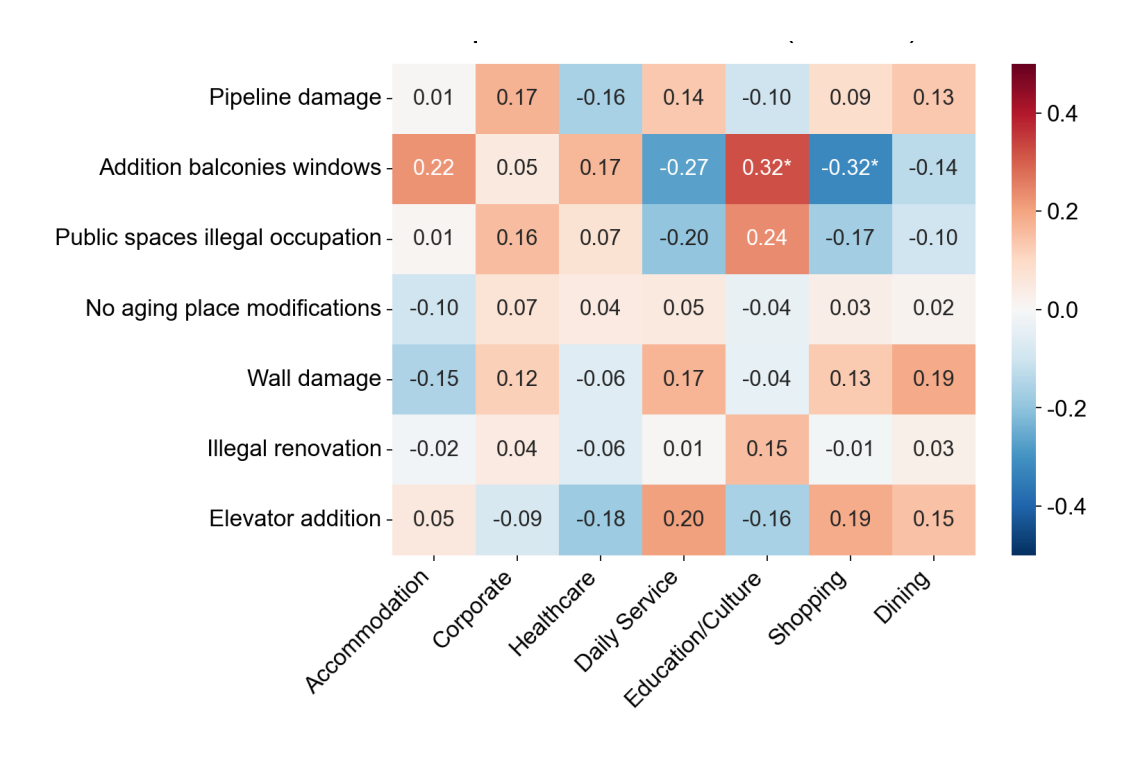}
    \caption{Spearman rank correlation (1000m)}
    \label{fig:sub_spearman_1000}
  \end{subfigure}

  \vspace{0.6em}

  \begin{subfigure}{0.48\textwidth}
    \centering
    \includegraphics[width=\linewidth]{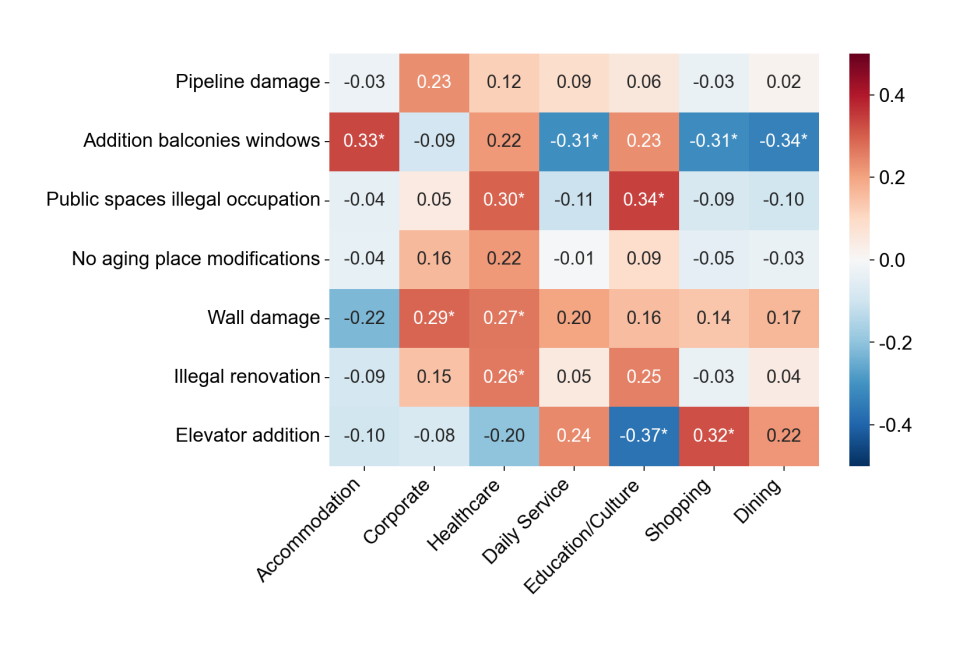}
    \caption{Spearman rank correlation (1500m)}
    \label{fig:sub_spearman_1500}
  \end{subfigure}

\caption{Correlation heatmaps of multi-scale POI fusion post-correction experiments.}
\label{fig:correlation_total}
\end{figure*}

After completing micro-scale visual perception and macro-scale community functional coupling, this section conducts post-correction on the judgment of building health check-up issues in old communities through the MSCF framework.
Based on the statistical association mechanism established in Section 3.3.2, environmental prior information under different POI radii (500m, 1000m, 1500m) is input into the classifier.
The Pearson linear correlation heatmaps and Spearman rank correlation heatmaps between the proportion of POI features and the incidence rate of physical examination issues under different spatial radii are shown in Figures~\ref{fig:sub_pearson_1500},  \ref{fig:sub_spearman_1000}, and \ref{fig:sub_spearman_1500}, respectively.
Among them, the Pearson linear correlations at the 500m and 1000m scales, as well as the Spearman rank correlation at the 500m scale, are omitted here because no significant correlated pairs emerged after FDR correction.
The correlation analysis was used as a spatial plausibility test rather than as direct evidence of predictive superiority.
It indicates that several physical examination issues are not randomly distributed with respect to surrounding urban functions, but show scale-dependent associations with community functional context.
In this study, POI features therefore represent the functional environment of old residential communities, including commercial activity, public service concentration, daily service intensity, and infrastructure exposure.
These contextual factors provide an interpretable spatial prior for building-level inspection, but their predictive value must still be verified through out-of-sample ablation experiments.

As can be seen from Figure~\ref{fig:sub_pearson_1500}, the 1000m scale in the Spearman rank correlation begins to capture the significant monotonic correlations between balcony addition and the proportion of POIs such as education/culture and shopping.
Figures~\ref{fig:sub_spearman_1000} and \ref{fig:sub_spearman_1500} show that the 1500m scale exhibits the richest significant correlation matrix, particularly establishing statistical associations between elevator addition, public space occupation, and accommodation functions.
These results support the spatial relevance of POI-derived functional context, but they do not by themselves establish stable building-level predictive gain.
The added value of POI features is therefore evaluated separately by comparing MVVA and MSCF under community-isolated cross-validation.

Based on the above spatial environmental prior features, the post-correction judgment performance of different visual object detection front-ends under 500m, 1000m, and 1500m radii was tested, and the experimental results are shown in Table 4-6.
The results show that the post-correction judgment performance did not reach its highest value at the 1500m scale, which had the most significant univariate correlated pairs, but reached the best-performing value at the 1000m block scale in this dataset.
This result indicates that community-scale association and building-level prediction should be distinguished: stronger POI correlations do not automatically produce better out-of-sample classification.
Among them, the decision combination using YOLOv8 as the front-end detector fused with a 1000m POI context (YOLOv8 + 1000m MSCF) recorded the highest overall building-level judgment score, with an average F1-score reaching 76.79\%.
\begin{table}[htbp]
\centering
\caption{Average F1 ranking of major models under MSCF conditions}
\label{tab:mscf_ranking}
\begin{tabularx}{\linewidth}{
  >{\raggedright\arraybackslash}X 
  >{\centering\arraybackslash}X 
  >{\centering\arraybackslash}X
}
\toprule
\textbf{Model} & \textbf{Radius} & \textbf{F1} \\
\midrule
\textbf{YOLOv8} & \textbf{1000} & \textbf{76.79\%} \\
\addlinespace
YOLOv9 (1280) & 1500 & 76.67\% \\
\addlinespace
YOLOv9 (1280) & 1000 & 75.92\% \\
\addlinespace
YOLOv10 & 1500 & 74.67\% \\
\addlinespace
YOLOv9 (1280) & 500 & 74.52\% \\
\addlinespace
YOLOv8 & 500 & 74.45\% \\
\addlinespace
YOLO11 & 1000 & 74.27\% \\
\addlinespace
YOLOv8 & 1500 & 74.24\% \\
\addlinespace
YOLO11 & 1500 & 74.21\% \\
\addlinespace
RT-DETR & 1500 & 74.08\% \\
\bottomrule
\end{tabularx}
\end{table}
\begin{figure}
	\centering
		\includegraphics[scale=.9]{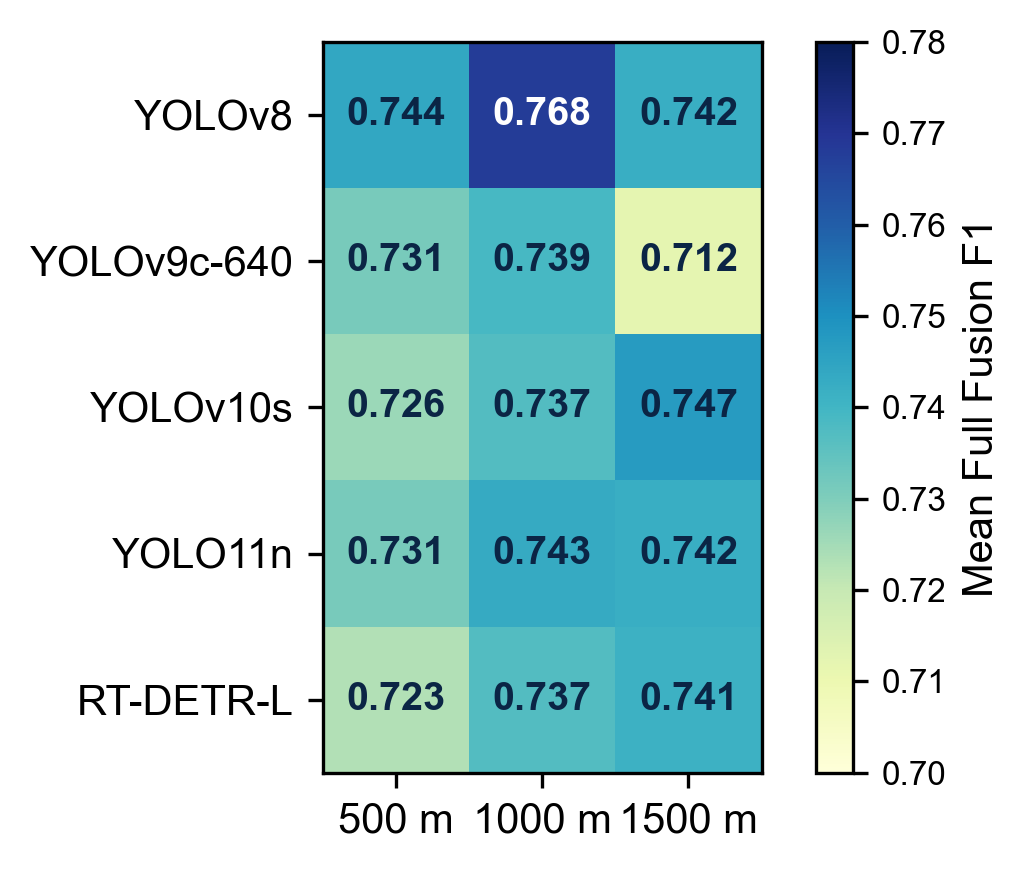}
	\caption{Average MSCF F1-score under different visual front-ends and POI radii.}
	\label{fig:4-5-redius}
\end{figure}

As shown in Section~\ref{sec4.3}, YOLOv9c-640 achieved the best image-level mAP among the tested detectors.
However, in the final multi-source building-level decision task, YOLOv8 achieved the highest F1-score (76.79\%) at the 1000m POI scale, whereas YOLOv9c-640 reached only 73.85\% under the same setting.
This discrepancy indicates that the best image-level detector is not necessarily the best front-end for building-level diagnosis.
Image-level mAP mainly evaluates box localization and classification in individual images, whereas building-level diagnosis depends on the stability and separability of aggregated features across all images of the same building.
These aggregated features include box counts, confidence distributions, infection rates, and their interaction with POI-derived contextual variables.
Therefore, a detector with higher image-level mAP may not produce the most discriminative building-level representation after multi-view aggregation.

Figure~\ref{fig:4-5-redius} reveals that as the POI Buffer Radius increases, the judgment performance of each model presents an inverted U-shaped characteristic of 'first rising, then falling'.
At the 500m neighborhood micro-scale, due to the spatial discreteness of the distribution density of commercial and public facility POIs around old communities, feature vectors are prone to generating statistical noise due to local 'zero values' or sparsity phenomena, failing to stably map the community functional background; whereas at the 1500m macro-scale, buffers easily cross urban main roads and mix with the external functional backgrounds of adjacent residential areas or commercial centers, resulting in a Spatial Smoothing Effect and Spatial Non-stationarity.
In comparison, the 1000m scale achieved the best-performing trade-off in this dataset between functional coverage breadth and spatial specificity, enabling the Random Forest classifier to capture the non-linear interaction between visual evidence and environmental priors.

This result supports the need to select the visual front-end according to the final building-level diagnostic objective rather than relying solely on image-level detection benchmarks.
For the refined governance of urban health check-ups, model selection should be guided by end-to-end building-level performance.
Therefore, this study determined YOLOv8 + 1000m as the main setting.
\subsection{Ablation Experiments and Class Difference Analysis}
\begin{figure}
	\centering
		\includegraphics[scale=.55]{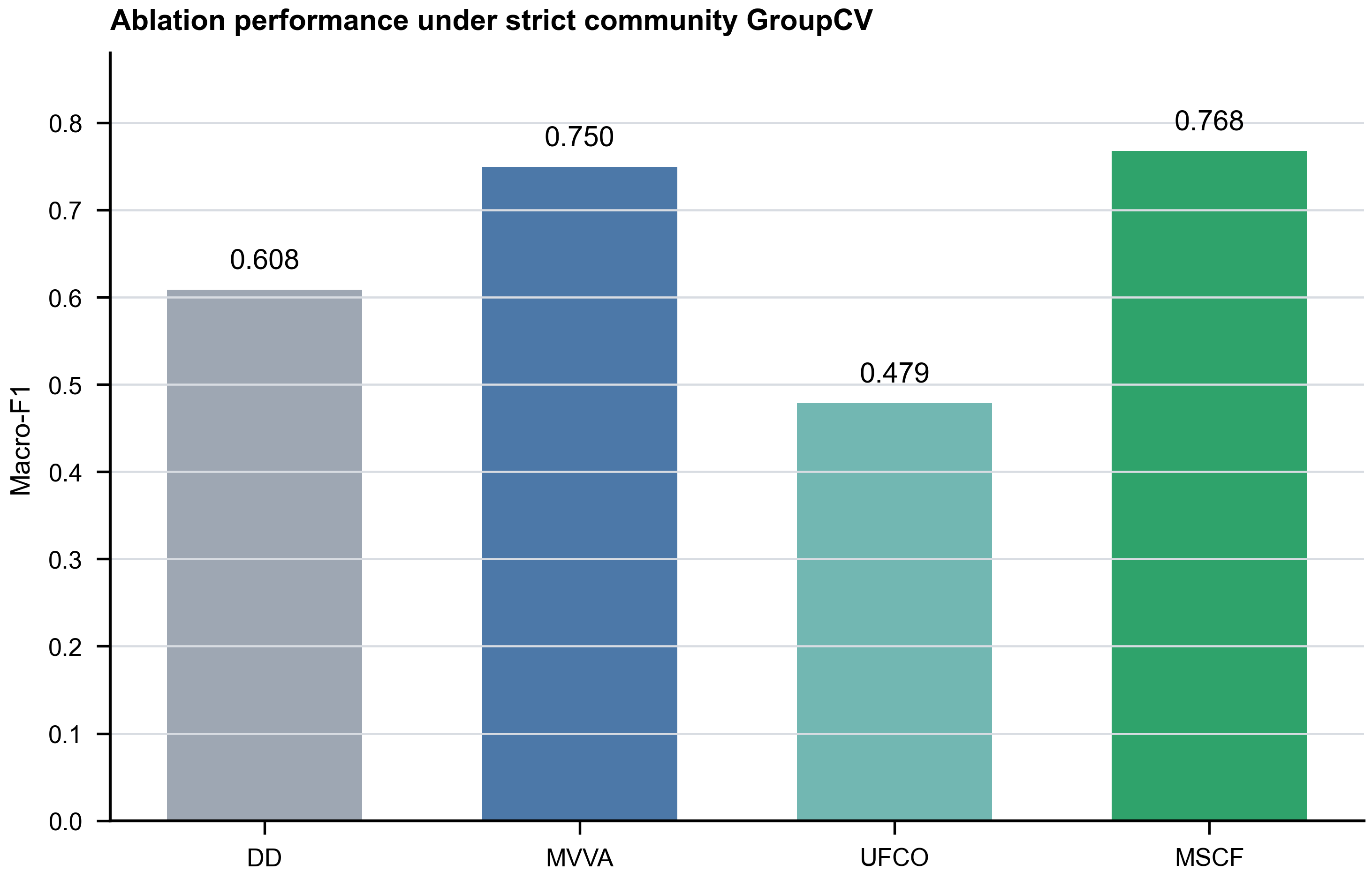}
	\caption{Average F1 of four ablation schemes under yolov8\_balanced + 1000m condition.}
	\label{fig:4-6-F1.png}
\end{figure}

To systematically verify the effectiveness and synergistic effects of each module in the proposed method, this section designed four sets of controlled ablation experiments.
The average F1 results of the four types of ablation experiments are shown in Figure~\ref{fig:4-6-F1.png}.
Considering that buildings within the same community share identical or highly similar POI contexts, normal random partitioning by building can easily trigger spatial data leakage, thereby overestimating the model's generalization capability.
Therefore, this study took strict Spatial Group CV as the main experimental protocol to ensure that buildings in the same community would not appear in the training and test sets simultaneously.
The specific settings for the four sets of experimental schemes are as follows:
For clarity, the four strategies are ordered from image-only to context-only and fused settings.
DD is the simplest image-triggered baseline, MVVA aggregates multi-view visual evidence, UFCO tests whether urban context alone is sufficient, and MSCF represents the proposed fusion strategy combining visual and POI features.

(1) Direct Detection (DD): If the object detection model detects a target box (confidence >= 0.15) in any image of a building, it directly judges that the building has this class of health check-up problem.

(2) Multi-View Visual Aggregation (MVVA): Instead of directly using detection boxes of single images as the final judgment, all multi-view images belonging to the same building are associated, and the 6-dimensional building-level soft visual statistical features extracted in Section 3.2.2 are utilized and input into a Random Forest classifier for secondary discrimination.

(3) Urban Functional Context Only (UFCO): Only the multidimensional community POI environmental features at the 1000m block scale (detailed in Section 3.3.1) are used as the input of the classifier, without relying on any image detection evidence.

(4) Multi-Source Context Fusion (MSCF): The core fusion scheme of this study described in Section 3.4. The proposed fusion scheme cascades building-level 6-dimensional visual statistical features and community POI environmental features into a joint feature vector (detailed in Section 3.4), uniformly inputting it to the post-correction classifier for comprehensive judgment.

\begin{figure}
	\centering
		\includegraphics[scale=.55]{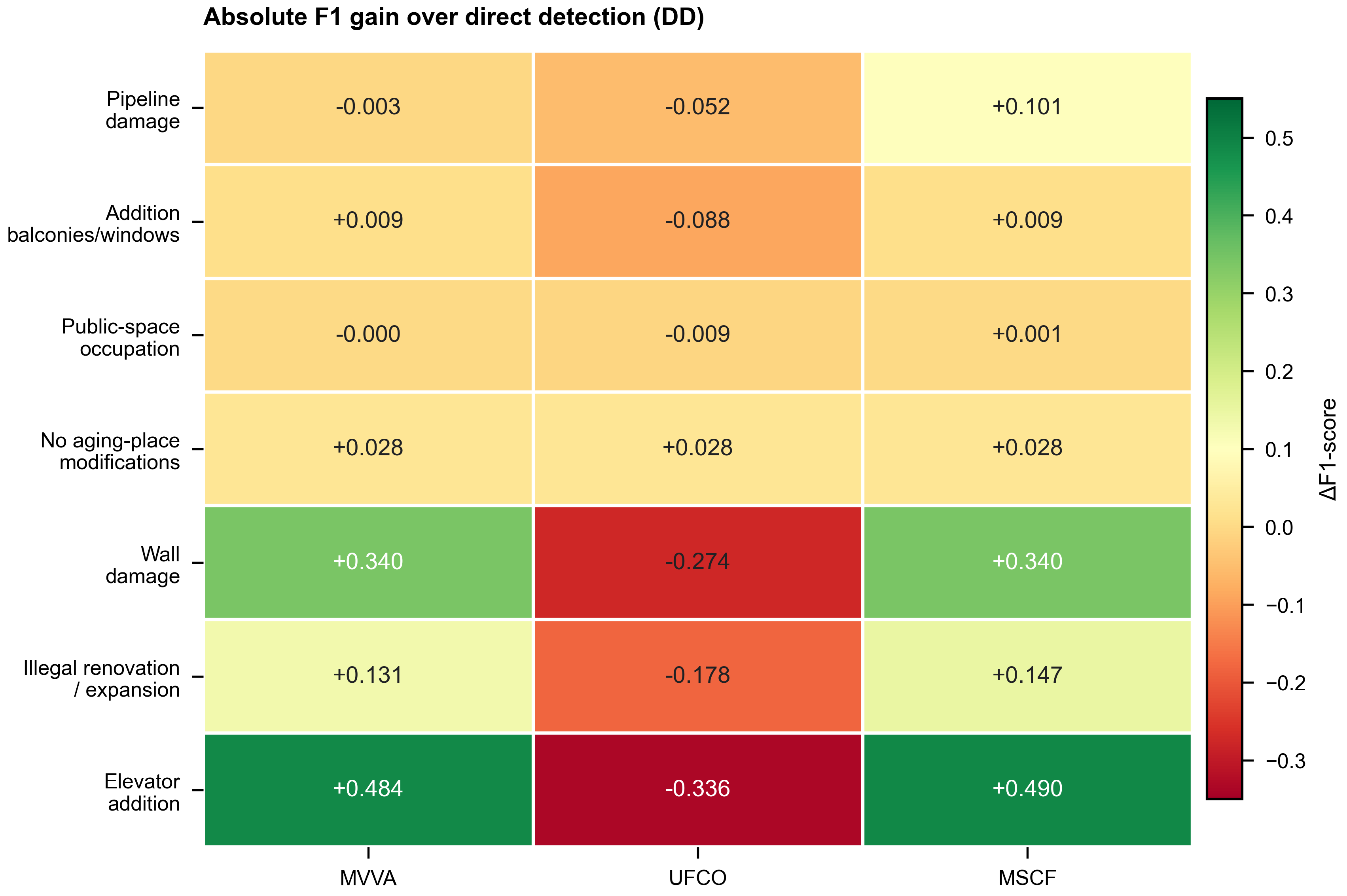}
	\caption{Heatmap of absolute F1-score gain variations based on the DD baseline}
	\label{fig:4-7-Heatmap}
\end{figure}

Because the formation mechanisms and external manifestations of various health check-up issues differ, the rectification gains brought by multi-source context fusion present obvious heterogeneity among different categories.
Table~\ref{tab:class_level_performance} and Figure~\ref{fig:4-7-Heatmap} comprehensively display the judgment efficacy of each scheme on the seven sub-categories of diseases, as well as the net performance gains relative to the DD baseline.
\begin{table*}[htbp]
\centering
\caption{Comparison of class-level building judgment performance under main settings}
\label{tab:class_level_performance}
\begin{tabular}{lcccccccc}
\toprule
\multirow{2}{*}{\textbf{Class}} & 
\multicolumn{2}{c}{\textbf{DD}} & 
\multicolumn{2}{c}{\textbf{MVVA}} & 
\multicolumn{2}{c}{\textbf{UFCO}} & 
\multicolumn{2}{c}{\textbf{MSCF}} \\
\cmidrule(lr){2-3} \cmidrule(lr){4-5} \cmidrule(lr){6-7} \cmidrule(lr){8-9}
& \textbf{F1} & \textbf{Acc} & \textbf{F1} & \textbf{Acc} & \textbf{F1} & \textbf{Acc} & \textbf{F1} & \textbf{Acc} \\
\midrule
Pipeline damage & 11.32\% & 94.09\% & 10.99\% & 97.45\% & 6.15\% & 94.24\% & 21.43\% & 99.31\% \\
\addlinespace
Addition balconies windows & 97.60\% & 96.10\% & 98.46\% & 97.55\% & 88.82\% & 79.89\% & 98.51\% & 97.64\% \\
\addlinespace
Public spaces illegal occupation & 94.93\% & 91.26\% & 94.91\% & 91.23\% & 94.04\% & 88.82\% & 95.03\% & 90.77\% \\
\addlinespace
No aging place modifications & 96.90\% & 94.03\% & 99.72\% & 99.44\% & 99.72\% & 99.44\% & 99.72\% & 99.44\% \\
\addlinespace
Wall damage & 42.61\% & 77.31\% & 76.60\% & 96.22\% & 15.25\% & 11.46\% & 76.60\% & 96.22\% \\
\addlinespace
Illegal renovation expansion & 38.98\% & 86.79\% & 52.05\% & 95.16\% & 21.18\% & 75.55\% & 53.69\% & 95.23\% \\
\addlinespace
Elevator addition & 43.57\% & 94.99\% & 91.94\% & 99.68\% & 10.00\% & 97.14\% & 92.56\% & 99.71\% \\
\midrule
\textbf{Mean (MACRO)} & \textbf{60.84\%} & \textbf{90.65\%} & \textbf{74.95\%} & \textbf{96.68\%} & \textbf{47.88\%} & \textbf{78.08\%} & \textbf{76.79\%} & \textbf{96.90\%} \\
\bottomrule
\end{tabular}
\vspace{1ex} \\
\end{table*}

Because several issue categories are highly sparse, overall Accuracy alone can be misleading.
In such cases, a classifier may obtain a very high Accuracy mainly by correctly identifying the large number of negative buildings, while still missing most positive cases.
Table~\ref{tab:sparse_class_confusion} therefore reports the valid sample size, positive rate, fold feasibility, and confusion matrix for the sparse classes under the primary community Group CV setting.
This table is used as a diagnostic aid for interpreting class imbalance, rather than as an additional model-ranking result.

\begin{table*}[htbp]
\centering
\caption{Sample size, fold feasibility, and confusion matrix for sparse classes under the primary community Group CV setting.}
\label{tab:sparse_class_confusion}
\scriptsize
\setlength{\tabcolsep}{3pt}
\begin{tabular}{lrrrrrrrrrrr}
\toprule
\textbf{Class} & \textbf{Valid N} & \textbf{Pos.} & \textbf{Pos. rate} & \textbf{Folds} & \textbf{Min test pos.} & \textbf{TP} & \textbf{FP} & \textbf{FN} & \textbf{TN} & \textbf{F1} & \textbf{Acc.} \\
\midrule
Pipeline damage & 3179 & 23  & 0.72\% & 3 & 6  & 3   & 2  & 20 & 3154 & 21.43\% & 99.31\% \\
Wall damage & 1745 & 154 & 8.83\% & 5 & 13 & 108 & 20 & 46 & 1571 & 76.60\% & 96.22\% \\
Illegal renovation expansion & 2892 & 152 & 5.26\% & 5 & 15 & 80  & 66 & 72 & 2674 & 53.69\% & 95.23\% \\
Elevator addition & 3151 & 63  & 2.00\% & 5 & 4  & 56  & 2  & 7  & 3086 & 92.56\% & 99.71\% \\
\bottomrule
\end{tabular}
\end{table*}

Synthesizing the experimental results, the post-correction effects of environmental features on different issues can be summarized into the following two core mechanisms.

The first category comprises those that may be affected by the macroscopic built-up area functional structure and obtain certain supplementary benefits with the assistance of the environmental context, typically including Pipeline Damage and No Aging Place Modifications. Among them, the baseline DD F1-score for pipeline damage was only 11.32\%;
after introducing the 1000m block life circle functional features, MSCF improved to 21.43\%, with an absolute net gain of +10.11\%.
However, this improvement should be interpreted conservatively.
As shown in Table~\ref{tab:sparse_class_confusion}, pipeline damage contained only 23 positive building-label pairs, accounting for 0.72\% of the valid samples.
Its Accuracy reached 99.31\%, but this value was mainly driven by 3154 true-negative buildings rather than by reliable positive detection.
Only 3 positive cases were correctly detected, while 20 were missed.
For sparse categories, Accuracy was therefore treated as a supplementary metric, while F1-score, Recall, and the underlying confusion matrix were used as the primary basis for interpreting recognition capability.
Because of this sparsity, pipeline damage was evaluated with 3-fold community Group CV, with at least six positive samples in each test fold and no train-test community overlap.
The other sparse categories retained 5-fold community Group CV, with minimum test-fold positive counts ranging from 4 to 15 and zero train-test community overlap.
Accordingly, the result for pipeline damage is best understood as an improvement in high-precision screening capability rather than as a complete solution to sparse category recognition.
No aging place modifications reached 99.72\% in both MVVA and MSCF, indicating that this category is mainly supported by stable visual aggregated features.
Addition Balconies Windows already reached 98.46\% in the MVVA stage, and MSCF was 98.51\%, which has very minimal additional improvement compared to visual aggregation, and more reflects a supplementary explanation of spatial distribution mechanisms.

The second category comprises those more reliant on fine visual soft features, mainly including Wall Damage, Illegal Renovation Expansion, and Elevator Addition. Wall damage was 76.60\% under both MVVA and MSCF, indicating that POIs did not provide extra discriminative gain, and the key improvement came from multi-view visual aggregation.
Illegal renovation expansion improved from 52.05\% in MVVA to 53.69\% in MSCF, a gain of +1.64\%;
elevator addition improved from 91.94\% to 92.56\%, a gain of +0.62\%.
Thus, the POI context mainly exerts a weak corrective effect on local component-type or strong visual feature categories and cannot replace image evidence.
The consistently weaker performance of UFCO is consistent with this boundary.
Because UFCO uses only POI-derived environmental features without visual evidence, its lower Macro-F1 indicates that surrounding urban functions alone are insufficient for reliable building-level issue recognition.
POI variables should therefore be interpreted as contextual priors rather than direct diagnostic evidence or causal determinants of building condition.
Their value lies in complementing multi-view visual aggregation in uncertain cases, improving probability ranking, and providing spatially interpretable context for renewal governance.

\begin{figure*}[b]
  \centering
  \includegraphics[width=0.9\textwidth]{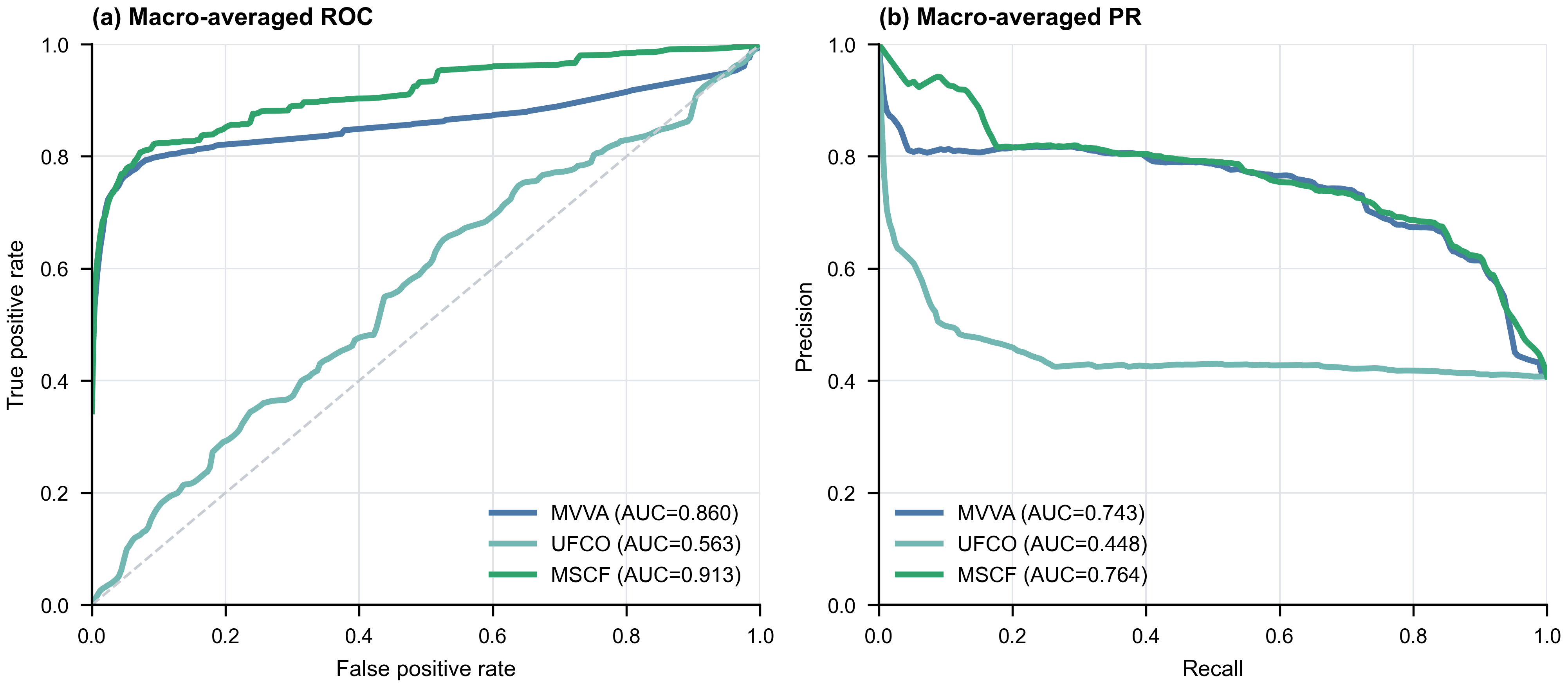} 
  \caption{Comparison of ROC and PR curves for major fusion models under multi-class decision-making.}
  \label{Fig4-8-PR.png}
\end{figure*}

From a global perspective of probabilistic prediction, Figure~\ref{Fig4-8-PR.png} displays the ROC and PR curves of the main ablation models under multi-category decision-making.
The ROC curves indicate that the MSCF framework with added POI information not only outperforms UFCO in global positive and negative sample discrimination, but also matches or slightly surpasses MVVA, indicating that multi-source feature fusion effectively improves the soft ranking capability of the model's output confidence.
However, the more stringent PR curves intuitively reveal the severe test posed to the model by extremely imbalanced samples.
On extremely rare long-tail diseases like Pipeline damage, the obvious fluctuations of the PR curves objectively reflect the inferential instability caused by data scarcity.
\subsection{Spatial Generalization and Robustness Analysis}
This subsection is designed as a robustness and boundary analysis rather than as the primary model-ranking experiment.
The main performance comparison has been reported in Section~\ref{Sec:POI fusion}; here, the purpose is to examine how the conclusions change under stricter spatial partitioning, limited POI gains, and different post-correction classifiers.
Accordingly, Normal CV is interpreted as an identically distributed upper-bound reference, community Group CV is used as the primary within-city generalization test, and subdistrict/district GroupCV is treated as a spatial stress test.

\subsubsection{Within-City Generalization in Cross-Community Scenarios}
As described in Section 4.5, to prevent the model from causing spatial data leakage by rote-memorizing environments of the same community, Spatial Group CV utilizes community-level group cross-validation taking 92 communities as grouping units to evaluate the model's within-city generalization capability on unseen communities. 
Traditional Normal CV can evaluate the model's average performance on identically distributed samples, but because the training and testing sets may share the POI context of the same community, its results are more suited as an upper bound for identically distributed performance, rather than primary evidence for spatial extrapolation. Under Normal CV, the macro-average F1 of the main fusion model inflated to 81.50\%;
whereas after implementing community isolation, the score receded to 76.79\% (Accuracy 96.90\%), reflecting within-city cross-community extrapolation under the Qingdao sample. After filtering out the inflated margin, this score still outperformed the DD baseline (F1=60.84\%, ACC=90.65\%) and the MVVA visual aggregation scheme (F1=74.95\%, ACC=96.68\%).

\begin{figure}
	\centering
		\includegraphics[scale=.5]{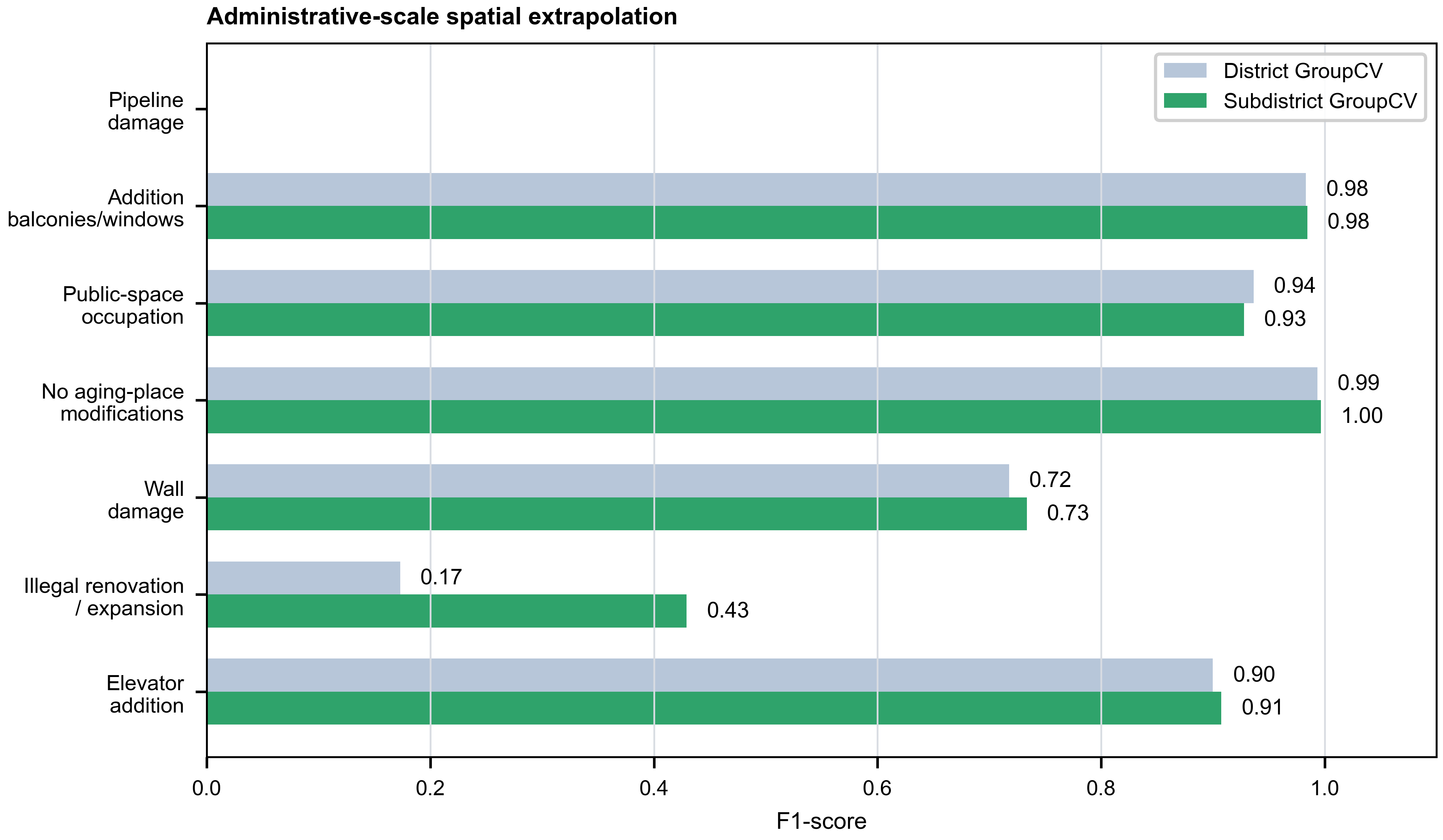}
	\caption{Class-level F1 under administrative-scale spatial extrapolation stress tests: Subdistrict GroupCV vs. District GroupCV}
	\label{fig:Fig4-9-CV}
\end{figure}

After the community-level evaluation, this study further conducted higher-difficulty cross-administrative isolation tests within Qingdao, taking subdistricts and districts as barriers, with results shown in Figure~\ref{fig:Fig4-9-CV}.
This experiment does not serve as the primary performance evaluation protocol, but is used to test the model's categorical stability under stronger spatial distribution shift conditions.
Subdistrict GroupCV requires the test subdistrict as a whole not to participate in training, and District GroupCV further requires the test district as a whole not to participate in training; hence, its difficulty is higher than community-level isolation.

As shown in Figure~\ref{fig:Fig4-9-CV}, with the increase of the test span, various issues exhibited different generalization resilience:
issues like "no aging place modifications" and "addition balconies windows" remained stable in cross-district tests, indicating that they have relatively transferable patterns across different urban sectors within Qingdao;
whereas "illegal renovation expansion" dropped sharply in cross-administrative district predictions, reflecting macroscopic differences in urban management law enforcement intensity across jurisdictions;
the extremely low-frequency "pipeline damage" directly dropped to zero, showing that extremely sparse samples cannot support cross-district predictions.
\subsubsection{Descriptive and Algorithmic Validation of POI Auxiliary Effects}
Because the macro-average F1 improvement of MSCF relative to MVVA is limited, it is still necessary to judge whether this improvement is consistent across detector and radius settings.
To keep the comparison directly traceable to the final community GroupCV summary, this paper first reports the paired macro-F1 differences between MSCF and MVVA under matched detector and POI-radius settings.
The resulting descriptive gains are shown in Table~\ref{tab:traceable_gains}.
\begin{table}[htbp]
\centering
\caption{Traceable macro-F1 gains of Multi-Source Context Fusion (MSCF) relative to Multi-View Visual Aggregation (MVVA)}
\label{tab:traceable_gains}
\setlength{\tabcolsep}{4pt} 
\small 
\begin{tabular}{lcccc}
\toprule
\textbf{Model} & \textbf{Radius} & \textbf{MVVA F1} & \textbf{MSCF F1} & \textbf{$\Delta$F1} \\
\midrule
RT-DETR-L   & 1000 & 73.15\% & 73.83\% & +0.68\% \\
\addlinespace
YOLOv8      & 1000 & 74.95\% & 76.79\% & +1.84\% \\
\addlinespace
YOLOv9c-1280& 1000 & 74.64\% & 75.92\% & +1.28\% \\
\addlinespace
YOLOv10s    & 1000 & 72.87\% & 73.99\% & +1.11\% \\
\addlinespace
YOLOv8      & 500  & 74.45\% & 74.45\% & +0.00\% \\
\bottomrule
\end{tabular}
\end{table}

Table~\ref{tab:traceable_gains} shows that the additional gain of MSCF over MVVA was positive but modest under the main community GroupCV setting.
For example, YOLOv8 + 1000m improved from 74.95\% to 76.79\%, corresponding to an absolute macro-F1 gain of +1.84 percentage points.
The near-zero gain of YOLOv8 + 500m also indicates that POI context does not uniformly improve all radius settings.
These values are reported as descriptive paired differences rather than formal significance tests.
The POI contribution is therefore interpreted as a limited and setting-dependent auxiliary correction to multi-view visual aggregation, rather than as a uniformly significant or decisive replacement signal.

Furthermore, to test whether the main conclusion depends on a specific post-corrector, this paper horizontally compared the performance of different machine learning classifiers under the same community GroupCV protocol and the same multi-source feature input, with results shown in Table~\ref{tab:fusion_classifiers}.
\begin{table}[htbp]
\centering
\caption{Average performance of different fusion classifiers under the primary setting}
\label{tab:fusion_classifiers}
\begin{tabular}{lccc}
\toprule
\textbf{Fusion Classifier} & \textbf{Precision} & \textbf{Recall} & \textbf{F1-Score} \\
\midrule
RandomForest & \textbf{83.64\%} & 74.56\% & \textbf{76.79\%} \\
\addlinespace
GradientBoosting & 77.49\% & 72.57\% & 74.71\% \\
\addlinespace
LinearSVM & 79.18\% & 74.32\% & 76.54\% \\
\addlinespace
LogisticRegression & 77.44\% & \textbf{78.04\%} & 76.55\% \\
\bottomrule
\end{tabular}
\end{table}

Table~\ref{tab:fusion_classifiers} shows that Random Forest achieved the highest macro-average F1 of 76.79\%.
LogisticRegression and LinearSVM reached comparable F1-scores of 76.55\% and 76.54\%, respectively, while GradientBoosting reached 74.71\%.
These results indicate that the multi-source feature fusion framework is not dependent on a single classifier family.
This paper ultimately selected Random Forest as the main classifier, primarily valuing its practical advantages such as requiring no complex parameter tuning and being capable of handling non-linear features.

\subsubsection{Spatial Diagnostic Maps and Urban Interpretation}
\begin{figure*}[b]
  \centering
  \includegraphics[width=\textwidth]{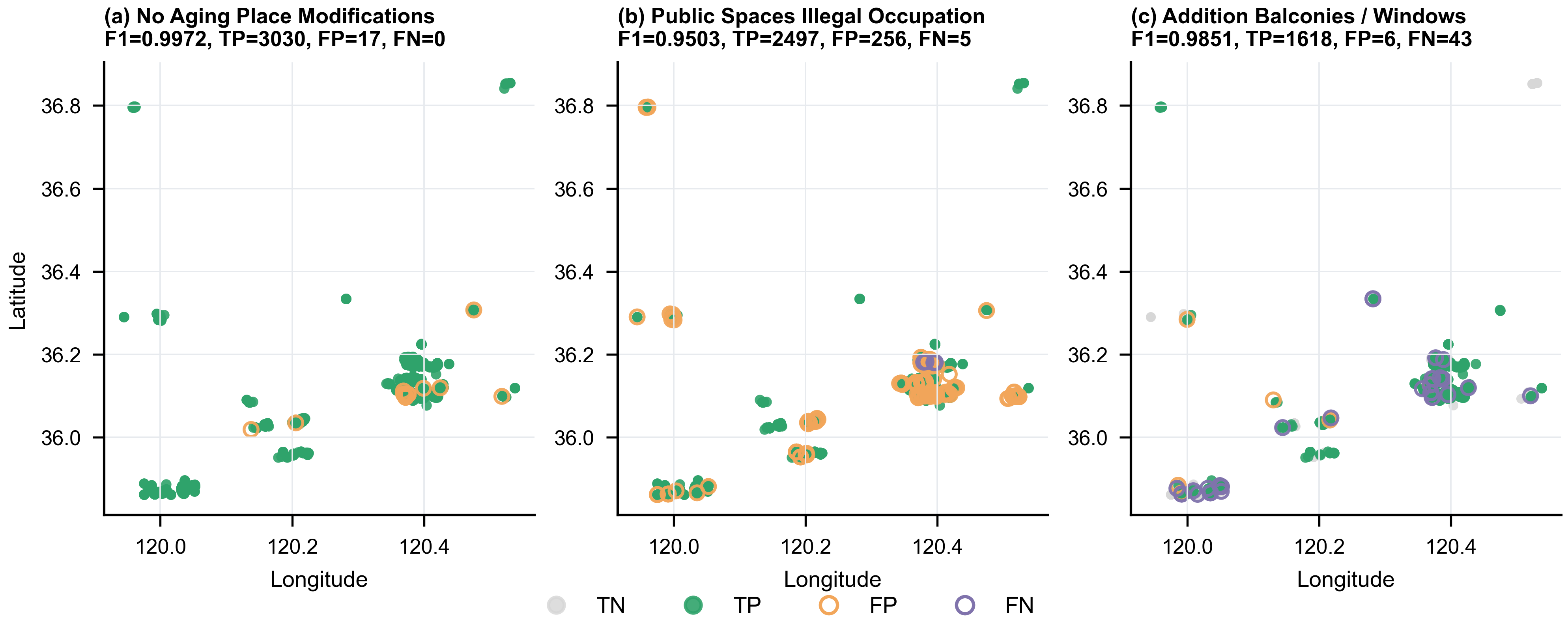} 
  \caption{Building-level spatial diagnostic maps for three types of stable health check-up issues.}
  \label{Fig4-10.png}
\end{figure*}

In addition to global numerical metrics, spatial diagnostics should place the Ground Truth labels, visual inspection results, and post-correction predictions into a unified spatial geographical framework for comparison, allowing spatial consistency and distribution topological features beyond the F1-score to be intuitively examined.
Such spatial review holds significant reference value for deploying deep learning models in urban management practices. By spatially linking each building ID with latitude and longitude coordinates, and based on the prediction status of the MSCF, building-level spatial distribution maps distinguishing True Negatives (TN), True Positives (TP), False Positives (FP), and False Negatives (FN) were drawn (Figure~\ref{Fig4-10.png}).
If the spatial agglomeration characteristics of the post-corrected positive buildings align closely with manual annotations, it indicates that the POI environmental context likely provided auxiliary rectification information; if obvious spatial misalignment still exists, it helps to diagnose the root causes of systemic misjudgments under specific geographical contexts.
This spatial diagnostic result can serve as a "decision map" reference in urban renewal governance: urban management departments can accordingly identify spatial units where FP and FN are relatively concentrated, providing geographical clues for subsequent manual review, on-site patrols by grid workers, and the prioritization of remediation.

From the spatial distribution morphology of the three types of stable health check-up issues in Figure~\ref{Fig4-10.png}, it can be seen that different disease categories present distinct spatial error topologies.
First, no aging place modifications (F1 = 99.72\%) exhibits strong global spatial homogeneity, with its TPs distributed in large contiguous areas and only a few FP points.
This reflects that the absence of aging-friendly facilities has strong systemic characteristics within the Qingdao old community samples. Second, public spaces illegal occupation (F1 = 95.03\%) demonstrates local agglomeration, with its FP points mostly sparsely distributed at the boundary zones between old communities and surrounding commercially active blocks.
This may occur because the peripheries of buildings near commercial blocks often have tables, chairs, or awnings compliantly placed by merchants, causing semantic confusion for the visual models, while POI features failed to completely isolate them. Lastly, the FN points for addition balconies windows (F1 = 98.51\%) aggregate locally in the old urban area of Shibei District, where building density is high and alleys are narrow.
This implies that ground visual collection faces constraints such as shadow occlusion and line-of-sight blockage under high-density urban textures, and subsequent supplementation through more comprehensive viewpoint collection or manual review is still required.

\begin{figure*}[b]
  \centering
  \includegraphics[width=0.9\textwidth]{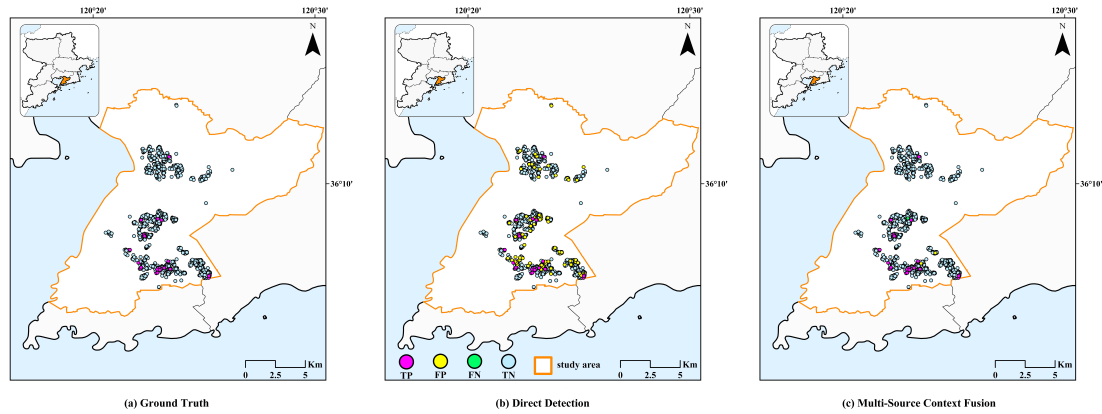} 
  \caption{Elevator addition local spatial diagnostic triptych.}
  \label{Fig4-11.png}
\end{figure*}

To further dissect the spatial distribution and error mechanisms of key livelihood renovation projects, this study generated a locally magnified spatial diagnostic triptych for elevator addition (Figure~\ref{Fig4-11.png}).
From the middle map (model prediction distribution) and the left map (ground truth label distribution) in Figure~\ref{Fig4-11.png}, it can be observed that the TP predicted by the model exhibit strong clustered agglomeration, restoring the true spatial evolution morphology of urban grid-based promotion of old community renovations.
Meanwhile, in the spatial diagnosis on the right map, FP and FN are significantly compressed, appearing only as sporadically distributed isolated scatterings, without large-scale, systemic regional spatial misjudgments.

\section{Discussion}
\subsection{Compensatory Mechanisms and Extrapolation Rigor of Multi-Scale Spatial Contexts}
The MSCF framework proposed in this study establishes an associative correction path between "microscopic physical disease perception" and "macroscopic urban functional context" at the methodological level.
This study does not regard POIs as direct "physical causal variables" of housing-related issues, but rather as spatial proxy variables reflecting community functional structure, maintenance pressure, and policy governance environment.
The research results emphasize that when constructing an urban-level health check-up system, the evaluation and selection of algorithms should not in isolation pursue the detection accuracy of single images, but should take the end-to-end building-level final decision performance as the core orientation.
In the Qingdao old community samples, the 1000m Buffer Radius provided the best observed balance between functional context coverage and spatial specificity.
Compared with the 1500m scale, this radius reduced the excessive smoothing caused by heterogeneous functions from adjacent areas.
Furthermore, strict community isolation validation (Spatial Group CV) showed that MSCF retained a modest performance advantage after excluding spatial leakage from buildings in the same community.

For issue types with distinct external features, such as illegal balcony additions and elevator additions, adding POIs did not bring a substantial score increase even when spatial correlations were observed.
This pattern indicates that the predictive contribution of POI features is conditional on the residual uncertainty left by visual aggregation.
In the proposed fusion framework, the core application value of POIs is reflected in three aspects:
First, when issues are tiny, hidden, or visually obscured, POIs provide contextual priors that may help adjust missed detections and false alarms.
Second, even when they do not change the final binary label, POIs can affect probability ranking and support risk-based inspection prioritization.
Finally, POIs help relate detected issues to surrounding urban functions, providing spatially interpretable evidence for renewal governance without implying direct causality.
In summary, multi-view image features remain the primary source for improving recognition accuracy, while the macroscopic urban environment provides auxiliary and interpretable context when visual evidence is incomplete or ambiguous.

\subsection{Collaborative Decision-Making Value Oriented Towards Grid-based Renewal Governance}
In the practice of urban renewal and refined governance, pure machine vision recognition is highly susceptible to interference, while full-coverage manual dragnet health check-ups face high administrative costs.
The MSCF framework provides a "spatially reviewable" intelligent auxiliary decision-making paradigm for the above pain points by projecting complex algorithmic judgments back to spatial entities with precise geographic coordinates (such as the spatial diagnostic maps in Figure~\ref{Fig4-10.png} and Figure~\ref{Fig4-11.png}).
Management decision-makers can rely on this high-resolution probability map to identify high-incidence areas of True Positives, guiding the priority placement of special funds;
meanwhile, they can target regions where FP and FN are relatively concentrated as coordinates for precise on-site verification by grid workers.
This closed-loop decision-making model of "prediction map pre-screening — spatial error diagnosis — manual grid response" not only greatly improves the targeting of manual review, but also provides a strong digital foundation support for the optimal allocation of inspection resources.
\subsection{Limitations and Future Research Directions}
This study still has certain limitations in exploring multi-source data fusion diagnosis, which also points out directions for future research:
First is the diagnostic bottleneck for extremely sparse long-tail samples.
For highly sparse categories with extremely strong local spatial heterogeneity, such as Pipeline damage and Elevator addition, spatial extrapolation still faces the risk of generalization collapse.
In the future, there is an urgent need to introduce targeted few-shot Active Learning mechanisms or rely on long-term data backflow for targeted supplementary sampling to stabilize the evaluation variance of long-tail categories.
Second, the empirical validation was conducted within old residential communities in Qingdao.
Although the community-isolated Spatial Group CV reduced within-community spatial leakage and tested extrapolation to unseen communities within the same city, it does not constitute external validation across cities.
Urban morphology, construction periods, housing maintenance regimes, governance practices, and POI ecosystems may differ substantially among cities.
Therefore, the cross-city generalizability of the proposed framework remains to be tested using independent datasets from other urban contexts.
Future work should extend the evaluation to multi-city samples and examine whether the best-performing POI radius, feature importance patterns, and fusion gains remain stable under different regional planning and management systems.
Third is the multi-dimensional expansion of visual perception dimensions.
Current images mainly originate from ground-collected multi-view facade images;
for inspection issues such as blind spots on the back of buildings, hidden dangers inside corridors, or roof leaks, an effective perception path is still lacking.
Subsequent research needs to introduce finer-grained urban environmental variables to fill physical blind spots.
Fourth is the dynamicization of the diagnostic context and human-machine collaborative evolution.
Static POIs can only represent the solid-state configuration of urban functional elements, and cannot yet reflect the human vitality and subjective social perception changes of the community.
Future research can integrate dynamic spatiotemporal flow data such as mobile phone signaling to construct a multidimensional diagnostic context;
at the same time, real-time access of dynamic feedback data from on-site verification by patrol grid workers into the post-correction classifier can be used to establish an adaptively iterative "human-machine collaborative closed-loop governance mechanism."
\section{Conclusion}
Aiming at the challenges of large-scale diagnosis in old communities, high labor costs, and the susceptibility of single-source machine vision to local occlusion misjudgments, this paper proposed an old community building-level health check-up framework that integrates micro multi-view visual aggregation and macro multi-scale environmental functional context fusion.
Validation based on the Qingdao field health check-up dataset indicates that this framework demonstrates a certain degree of usability in building-level recognition, spatial review, and evaluation of unknown communities within the same city.
Specific conclusions are as follows:

(1) The constructed micro MVVA mechanism maps fragmented image-level object detection information into building-level spatial diagnostic features, effectively mitigating the geometric randomness and local occlusion bias of single image perspectives.

(2) Under the Qingdao old community samples and the current POI feature system, the 1000m Buffer Radius was the best-performing radius for representing the surrounding functional context in this dataset.
The post-correction classifier at this scale achieved a better balance between neighborhood functional correlation and regional heterogeneous functional noise.

(3) Under strict community isolation validation, MSCF reached a Macro-F1 of 76.79\%, outperforming the DD baseline (60.84\%) and the MVVA baseline (74.95\%).
This quantitative result shows that multi-view visual aggregation is the main source of performance improvement, while the macroscopic urban functional context can provide limited but interpretable auxiliary corrections on this basis.

(4) Class-level analysis indicates that the system performs stably on No aging place modifications (F1 = 99.72\%), Public spaces illegal occupation (F1 = 95.03\%), and Addition balconies windows (F1 = 98.51\%);
however, for sparse small-sample issues such as Pipeline damage and Elevator addition, spatial generalization bias still exists, relying on subsequent supplementary sampling or grid-based manual inspections to supplement the diagnosis.

(5) Building-level spatial diagnostic maps can project model outputs back to specific geographical locations, providing auxiliary spatial references for spatial review, inspection prioritization, and existing housing renewal decision-making.

\section*{CRediT authorship contribution statement}
Kun Zhao: Conceptualization, Methodology, Supervision, Funding acquisition, Writing--review \& editing. Helei Ren: Methodology, Software, Data curation, Formal analysis, Visualization, Writing--original draft. Guilin Tang: Data curation, Investigation, Validation, Writing--original draft, Writing--review \& editing. Tianyi Chen: Methodology, Urban planning interpretation, Writing--review \& editing. Zhehui Song: Data curation, Investigation, Visualization. Xing Liu: Software, Validation. Lijian Zhou: Supervision, Writing--review \& editing. Yuhong Zhao: Investigation, Urban spatial analysis. Xiang Gao: Software, Formal analysis. Jinming Jiang: Conceptualization, Resources, Writing--review \& editing. Qichao Ban: Conceptualization, Supervision, Project administration, Funding acquisition, Writing--review \& editing.

\section*{Declaration of Competing Interest}
The authors declare that they have no known competing financial interests or personal relationships that could have appeared to influence the work reported in this paper.

\section*{Data availability}
The dataset associated with this study is available from Science Data Bank: Qichao Ban, Kun Zhao, Tianyi Chen, et~al. (2026). HOUSED: Housing-dimensiOnal visUal inSpection imagE Dataset. V2. Science Data Bank. \url{https://doi.org/10.57760/sciencedb.28941} (accessed on 30 May 2026).

\section*{Code availability}
The public code package and anonymized tabular resources supporting the workflow are available at \url{https://github.com/RHL-123/ceus-vision-poi-fusion} (accessed on 22 July 2026).

\section*{Funding}
This work was supported by the Natural Science Foundation of Shandong Province (Grant Nos. ZR2025MS834 and ZR2023ME234) and the Humanities and Social Science Research Program of the Ministry of Education, China (Grant No. 24YJCZH115).

\bibliographystyle{cas-model2-names}

\bibliography{cas-refs}

\end{document}